\documentclass[sn-mathphys,Numbered]{sn-jnl}% Math and Physical Sciences Reference Style
\usepackage{graphicx}%
\usepackage{multirow}%
\usepackage{amsmath,amssymb,amsfonts}%
\usepackage{amsthm}%
\usepackage{mathrsfs}%
\usepackage[title]{appendix}%
\usepackage{xcolor}%
\usepackage{textcomp}%
\usepackage{manyfoot}%
\usepackage{booktabs}%
\usepackage{algorithm}%
\usepackage{algorithmicx}%
\usepackage{algpseudocode}%
\usepackage{listings}%
\usepackage{tabularx} % Smaller table fonts

\theoremstyle{thmstyleone}%
\theoremstyle{thmstyletwo}%

\theoremstyle{thmstylethree}%

\begin{document}

\title[Article Title]{Myanmar XNLI: Building a Dataset and Exploring Low-resource Approaches to Natural Language Inference with Myanmar}


\author*{\fnm{Aung Kyaw} \sur{Htet}*}\email{akhtet@gmail.com}

\author{\fnm{Mark} \sur{Dras}}\email{mark.dras@mq.edu.au}


\affil{\orgdiv{Department of Computing}, \orgname{Macquarie University}, %\orgaddres{\street{Street}, 
\city{Sydney}, 
%\postcode{100190}, 
%\state{State}, 
\country{Australia}}

%\affil[2]{\orgdiv{Department}, \orgname{Organization}, \orgaddress{\street{Street}, \city{City}, \postcode{10587}, \state{State}, \country{Country}}}

%\affil[3]{\orgdiv{Department}, \orgname{Organization}, \orgaddress{\street{Street}, \city{City}, \postcode{610101}, \state{State}, \country{Country}}}


\abstract{Despite dramatic recent progress in NLP, it is still a major challenge to apply Large Language Models (LLM) to low-resource languages.  This is made visible in benchmarks such as Cross-Lingual Natural Language Inference (XNLI), a key task that demonstrates cross-lingual capabilities of NLP systems across a set of 15 languages.

In this paper, we extend the XNLI task for one additional low-resource language, Myanmar, as a proxy challenge for broader low-resource languages, and make three core contributions.  First, we build a dataset called Myanmar XNLI (myXNLI) using community crowd-sourced methods, as an extension to the existing XNLI corpus.  This involves a two-stage process of community-based construction followed by expert verification; through an analysis, we demonstrate and quantify the value of the expert verification stage in the context of community-based construction for low-resource languages. We make the myXNLI dataset available to the community for future research. Second, we carry out evaluations of recent multilingual language models on the myXNLI benchmark, as well as explore data-augmentation methods to improve model performance. 
Our data-augmentation methods improve model accuracy by up to 2 percentage points for Myanmar, while uplifting other languages at the same time.
Third, we investigate how well these data-augmentation methods generalise to other low-resource languages in the XNLI dataset.
}




\maketitle

\section{Introduction}\label{sec1}

Recent advances in Large Language Models (LLM), most prominently demonstrated by ChatGPT,\footnote{\url{https://chat.openai.com}} have uplifted the overall capabilities of Natural Language Processing.
But even with the general progress, there are additional challenges in working with low-resource languages, critical for supporting minority communities with relatively slow digital adoption.
Typical challenges with such languages include lack of datasets and established benchmarks on NLP performance, and language specific problems that require custom solutions.
It can be challenging to create datasets in low-resource languages, due to limited content on the internet, and the lack of access to individuals with adequate skills and knowledge.
Consequently, existing NLP solutions often do not extend well for low-resource languages, with such languages significantly left behind in performance benchmarks or lacking benchmarks entirely.

Natural Language Processing for Myanmar language faces such low-resource challenges, only exacerbated by socioeconomic factors of its native country Myanmar, formerly known as Burma.
Myanmar language, also known as Burmese,\footnote{Both Burmese and Myanmar refers to the same language, but for consistency we will use the name Myanmar, as in the Unicode Standard:  \url {https://www.unicode.org/charts/PDF/U1000.pdf}} is a language with relatively slow digital adoption, with its online presence only recently boosted by the mass adoption of social media and the affordability of mobile phones and the internet at the grassroots.
Being a non-Roman script language, the Myanmar script is not supported by the ASCII standard, and while the Unicode Standard for Myanmar was eventually developed, ad-hoc encoding standards for Myanmar have emerged and been adopted widely in the interim.
Myanmar content on the internet thus varies in quality significantly, and very few good datasets in Myanmar are available to the community. 
A rapid increase in online content fueled by adoption of Social Media, combined with non-standard encoding variants and limited NLP capabilities, means that there is much left to be desired in the state of NLP for Myanmar language compared to more commonly used languages.

In the field of NLP more generally, Large Language Models (LLMs) pre-trained on massive amount of raw data from the internet are becoming more ubiquitous.
Typically based on Transformer architectures, pre-trained large language models such as BERT \citep{devlin-etal-2019-bert} and GPT-3 \citep{gpt3_NEURIPS2020_1457c0d6} have often outperformed other NLP approaches, achieving the state-of-the-art across many benchmarks. 
Usually starting with monolingual data such as English, these methods are increasingly extended to multilingual settings, using multilingual training data and performing natural language tasks for multiple languages.
Multilingual LLMs trained on more than a hundred languages have emerged and several cross-lingual benchmarks have been established.
Furthermore, multilingual models provide a promising solution for low-resource languages by means of cross-lingual transfer, in which learning from high-resource languages can be transferred towards similar tasks in low-resource languages.

However, such emerging multilingual methods have not yet been explored widely in the context of Myanmar language.
Although many applications including mass internet platforms are increasingly becoming multilingual, there is limited previous work that involves Myanmar in multilingual settings. 
The Myanmar language can in fact provide interesting challenges for current multilingual models due to its linguistic characteristics.
For example, unlike many other languages with non-Roman scripts that have been the focus of NLP research, transliteration into and out of Myanmar is much less standardised, and certain Myanmar words may be written in multiple forms, posing generalisation challenges for the current multilingual models.
The Myanmar language itself has a particularly rich nominal morphology, and a rich numeral classifier system of the sort that is largely absent outside of East and South-East Asia.
%\footnote{World Atlas of Language Structures: https://wals.info/languoid/lect/wals_code_brm}
Focused research in this direction could thus not only improve the current state of NLP for Myanmar language, but also potentially provide challenges for existing multilingual frameworks, and further provide insights for other low-resource languages, towards making more robust and inclusive NLP systems.
In this paper, we consider one particular task and corresponding resources as a step in that direction.

Natural Language Inference (NLI), also known as Recognizing Textual Entailment (RTE), is an NLP task that requires recognising whether there is a logical entailment or contradiction between two natural language statements, or the lack thereof.
To correctly determine such logical relationships generally requires deep understanding of the semantics and context, therefore NLI is considered to be a central task for Natural Language Understanding.
In fact, early work on NLI such as \citet{williams-etal-2018-broad} argues that understanding entailment and contradiction is an important aspect for constructing semantic representations.
As NLP applications become increasingly multilingual, the NLI and other NLU tasks are also getting extended into multilingual settings.
Demonstrating the ability to reason in multiple languages or even across languages would indicate deeper levels of natural language understanding and semantic representations which are language-agnostic.
One canonical benchmark used to evaluate such cross-lingual NLI capabilities is Cross-lingual Natural Language Inference corpus (XNLI) \citep{conneau-etal-2018-xnli}.
The XNLI corpus provides the NLI benchmarking data in 15 different languages, covering a variety of language families from high to low-resource, and serves as a key evaluation benchmark for Cross-lingual Language Understanding (XLU).

Several multilingual models have been evaluated on the XNLI benchmark since its inception.
The XNLI authors established the very first XNLI performance scores by evaluating multilingual sentence encoders using BiLSTMs on the benchmark.  
Among subsequent models that obtained better XNLI performance, a cross-lingual language model XLM-R \citep{conneau-etal-2020-unsupervised} achieved remarkable improvements by pre-training on multilingual data on a large scale.
More recently, XLM-R is outperformed by mDeBERTa \citep{he2021debertav3} and became the state-of-the-art in XNLI.
But even though newer models have attained higher performance than their predecessors generally, there is usually a considerable gap in performance between high-resource and low-resource languages.
More research is thus required to uplift the XNLI performance of low-resource languages towards creating more inclusive language models.
And specifically for our context, XNLI does not contain a subcorpus for Myanmar language.

Hence the goal of our research is to explore the performance of state-of-the-art language models for Myanmar as a low-resource language, establish initial performance baselines in XNLI and find strategies to improve their performance.
To make our work applicable to not just Myanmar but other low-resource languages, our efforts focus on multilingual language models over monolingual language models. 
More specifically, our contributions in this paper are as follows: 

\begin{enumerate}
    \item We developed a dataset to benchmark the Natural Language Inference (NLI) performance in Myanmar language, namely Myanmar XNLI (myXNLI), by extending the existing XNLI dataset with Myanmar language counterparts to obtain training, validation and test datasets in Myanmar, as well as a parallel corpus joining Myanmar with the existing 15 XNLI languages.
    
    \item We used the myXNLI dataset to fine-tune a number of language models on the NLI task and evaluated them to establish the performance baselines for Myanmar language.
    The models in our baselines include multilingual models XLM-R \citep{conneau-etal-2020-unsupervised} and mDeBERTa \citep{he2021debertav3}, their monolingual counterparts RoBERTa \citep{roberta-zhuang-etal-2021} and DeBERTav3 \citep{he2021debertav3} respectively, and a monolingual Myanmar model MyanBERTa \citep{myanberta_ucsy_2022}.
    To the best of our knowledge, these baselines are the very first NLI benchmarks for Myanmar.
    \item  We examined which aspects of the process of constructing the dataset are important for improving performance.
    We also explored various data augmentation methods --- the exploitation of metadata such as Genre%, and multitask learning 
    --- designed to improve low-resource language performance, and showed that the maximum improvement over fine-tuning considering all of these methods individually or in combination is around the same as the improvement from fixing data quality.  
    \item We additionally evaluate our improvement methods against two reference low-resource languages, Swahili and Urdu.
    We analysed our results and present our view that these methods can be, in fact, useful for other low-resource languages. 
\end{enumerate}

%The Introduction section, of referenced text \cite{bib1} expands on the background of the work (some overlap with the Abstract is acceptable). The introduction should not include subheadings.

%Springer Nature does not impose a strict layout as standard however authors are advised to check the individual requirements for the journal they are planning to submit to as there may be journal-level preferences. When preparing your text please also be aware that some stylistic choices are not supported in full text XML (publication version), including coloured font. These will not be replicated in the typeset article if it is accepted. 

\section{Related Work}\label{sec2}

In this section we review the NLI tasks and datasets that we situate our new dataset with respect to, followed by the multilingual and Myanmar-language LLMs that we use for benchmarking performance on our new dataset.

\subsection{NLI, XLU and XNLI}
\label{sec:lit-rev-nli}

%\MDnote{230630}{I'm thinking it might be better to have this as the first section in this chapter.  For the Transformer section, to explain which models are best, you're referring to results on the XNLI dataset, whereas this one is relatively standalone.  It would also fit better with the revised RQs.}

\paragraph{Natural Language Inference (NLI)} 
NLI is a task that requires recognising whether there is a logical entailment, contradiction or neutrality between two different statements.
Given a pair of sentences \textit{Premise} and \textit{Hypothesis}, the goal of the task is to determine whether they are in an entailment relationship, or contradiction or otherwise neutral.
Based on the nature of the statements, the NLI task can impose challenges with varying levels of difficulties.
The entailment between \textit{Premise} and \textit{Hypothesis} is unidirectional and does not necessarily mean semantic equivalence or paraphrasing.
Entailment could encompass complex semantic relationships such as hierarchical (i.e. \textit{I like soccer} entails \textit{I like sports} but not necessarily vice versa) or commonsense knowledge (i.e. \textit{It is raining} entails  \textit{You need an umbrella}).
Other variations of NLI task may also exist, such as classification between \textit{Entailment} and \textit{Not Entailment} only. % as in where?
One of the largest NLI datasets in English is MultiNLI corpus \citep{williams-etal-2018-broad} which contains training, development and test datasets of 433k NLI sentence pairs in total across 10 genres. 
An example of NLI task in English is shown in Table~\ref{tab:nli-English} using MultiNLI sentence pairs, covering all three labels.

\begin{table}[h!]
        \begin{tabular}{l|l|l}
            \textbf{Premise} & \textbf{Hypothesis} & \textbf{Label}\\
            \hline            
            You don't have to stay there. & You can leave. & Entailment\\
            \hline
            You don't have to stay there. & You can go home if you want to. & Neutral
\\    
\hline
            You don't have to stay there. & You need to stay in that exact spot!
 & Contradiction
\\    
        \end{tabular}
        \caption{Example NLI Task in English using MultiNLI sentences}
        \label{tab:nli-English}
\end{table}

\paragraph{XNLI}
\label{ssec:lit-rev-xnli}
The Cross-Lingual Natural Language Inference (XNLI) dataset \citep{conneau-etal-2018-xnli} is a canonical benchmark used to evaluate combined NLI and cross-lingual understanding (XLU) capabilities.
The underlying dataset for XNLI mainly consists of development and test data in NLI 3-way format across 15 languages as a parallel corpus.
The range of the languages cover different language families as well as high and low-resource languages, making it an ideal evaluation benchmark for XLU.
Using crowd-sourcing methods, the core English portion of XNLI was constructed by sampling 250 sentences each from 10 text sources covering across a range of genres such as Government, Letters, Telephone, Travel and Fiction.
For each English source statement sampled as a premise, 3 hypotheses were manually generated, creating a total of 7500 human annotated development and test examples in NLI three-way classification format.
These premise-hypothesis pairs were manually labeled by 5 different annotators as \textit{entailment}, \textit{neutral} or \textit{contradiction}. 
Since different annotators may label a pair differently, a gold label was assigned based on the majority vote between 5 annotators.
The English portion was then translated by professional translators into 14 other languages --- French, Spanish, German, Greek, Bulgarian, Russian, Turkish, Arabic, Vietnamese, Thai, Chinese, Hindi, Swahili and Urdu --- making a total of 112,500 annotated examples.
The labels originally annotated for English pairs were also reused for the corresponding translated pairs for other languages.
% \MDnote{230725}{We probably want to emphasise that for community-led efforts like your Myanmar one this isn't necessarily the case, which is why we have a look at the effect of a second phrase of dataset construction.}
An example of XNLI task in two languages is shown in Table~\ref{tab:nli-French} as parallel sentences for English and French sharing the same labels.
In addition to the dev/test set, XNLI also includes a supplementary training dataset in the same 15 languages to enable training XNLI classifiers.
The authors of XNLI reused the MultiNLI corpus \citep{williams-etal-2018-broad} as the English portion of this training data, and used machine translation to create parallel training data in the 14 other languages.
While this training data is not part of the benchmark itself, it has proven to be useful for training multilingual models.
In fact, the authors showed that parallel data can help align sentence encoders in multiple languages, allowing classifiers trained in English to be reused towards other languages.
%XNLI can also act as a parallel data corpus across all 15 languages, as each premise or hypothesis sentence is translated into other 14 languages.

\begin{table}[h!]
%    \begin{center}
        \footnotesize
        \begin{tabular}{l|l|l}
            \textbf{Premise} & \textbf{Hypothesis} & \textbf{Label}\\
            \hline        
            You don't have to stay there. & You can leave. & \multirow{2}{*}{Entailment}\\
            Vous n'avez pas à rester là. & Tu peux partir.\\
            \hline
            You don't have to stay there. & You can go home if you want to. & \multirow{2}{*}{Neutral}\\
            Vous n'avez pas à rester là. & Vous pouvez rentrer à la maison si vous le souhaitez.\\    
\hline
            You don't have to stay there. & You need to stay in that exact spot!
 & \multirow{2}{*}{Contradiction}\\
            Vous n'avez pas à rester là. & Vous devez rester à cet endroit précis !
 &\\    
        \end{tabular}
        \caption{Example NLI Task in English and French using XNLI sentences}
        \label{tab:nli-French}
%    \end{center}
\end{table}

\paragraph{XTREME}
To meet the sophisticated demands of modern applications, NLU systems must aspire to support multiple natural language tasks rather than limited to a single particular task.
The General Language Understand Evaluation (GLUE) \citep{wang-etal-2018-glue} and SuperGLUE \citep{superglue_NEURIPS2019} benchmarks are created with this goal in mind, allowing a single model to be evaluated against multiple, well-established, core NLU tasks and compare with other models.
%These benchmark does not place constraints on the model architecture, except the requirement to accept single sentences or sentence-pairs in the input, and produce predictions in output according to the tasks.
While GLUE and SuperGLUE are English-only benchmarks, XTREME / XTREME-R \citep{ruder-etal-2021-xtreme} is a multilingual benchmark for evaluating cross-lingual generalisation across multiple NLU tasks.
The benchmark covers 50 languages from 12 typologically diverse language families, and its task categories include classification, structured prediction, question answering and retrieval.
As with the monolingual (English) benchmarks, Natural Language Inference is a significant aspect of the benchmark, and represented by the XNLI \citep{conneau-etal-2018-xnli} task and dataset. 
XTREME focuses on zero-shot cross-lingual transfer, where models can be pre-trained on any multilingual corpus but fine-tuned only in English, and evaluated against the benchmark.
The baselines established by XTREME leaderboard showed that, despite recent overall progress in NLU, there are still significant gaps in performance between high-resource and low-resource languages.

% Skipped next section
% \subsection{Observations on NLI Benchmarks} 

\subsection{The Emergence of Multilingual Models}

\subsubsection{XLM}

\citet{xlm_NEURIPS2019} showed in XLM that cross-lingual pre-trained language models possess improved performance on cross-lingual language tasks.
Their XLM approach extended BERT's Masked Language Modelling (MLM) objective into Translation Language Modelling (TLM) objective by using parallel sentences in different languages.
For example, to predict a masked English word using TLM, XLM can attend to both the English sentence and its French translation, encouraging the representations to align. 
This improved the performance over multiple cross-lingual tasks such as Cross-lingual Natural Language Inference (XNLI), Unsupervised Neural Machine Translation and Supervised Machine Translation.
They also demonstrated at the same time that low-resource languages can benefit from datasets in higher resource languages which share similar scripts.
As a specific example, the perplexity of a Nepali language model was reduced when trained additionally on datasets in Hindi which shared a similar script, together with English.

%\subsubsection{RoBERTa}

%On monolingual model approaches,  evaluated several design decisions in pre-training BERT models and combined their best approaches in RoBERTa.
%Precisely, they found that pre-training longer with bigger batches over more data, with longer sequences, and using dynamic masking pattern for Masked Language Modelling (MLM) has significantly improved the model performance over several downstream tasks, while the next sentence prediction (NSP) objective is no longer necessary to match earlier performance in BERT.
%Based on the BERT architecture, RoBERTa implemented the above configurations and achieved state-of-the-art results on multiple benchmarks including GLUE.
% \MDnote{230630}{You mention these benchmarks and several times, so I think you need to explain them briefly. A good place would probably be at the end of the Transformer section: You've noted that there are a lot of models, and the benchmarks were developed to allow comparison.}

\subsubsection{XLM-R}

Building on top of mBERT, XLM and RoBERTa \citep{roberta-zhuang-etal-2021} approaches, \citet{conneau-etal-2020-unsupervised} showed the effectiveness of pre-training large scale cross-lingual language models through XLM-R.
Although based on mBERT architecture with a multilingual MLM pre-training objective, XLM-R used optimised design decisions and configurations suggested by RoBERTa.
It was also pre-trained on a multilingual corpus, using CommonCrawl (CC) Corpus\footnote{\url{https://commoncrawl.org}} with 100 languages (2.5TB). 
This is much larger than the datasets used by mBERT and XLM which are based on Wikipedia.
XLM-R demonstrated that pre-training multilingual models at large scale improved the downstream cross-lingual tasks, achieving significant improvements on several benchmarks including XNLI and GLUE.
In particular, it observed improved representation of low-resource languages, demonstrated by significant uplifts in XNLI performance on Swahili and Urdu.
XLM-R also exposed some properties and limitations of multilingual models.
For a fixed size model, as the number of languages increased in pre-training, per-language capacity decreases.
However, this can be alleviated by increasing the model size.
%Also, the performance on low-resource language can be improved by pre-training on a related higher-resource language data, but the overall downstream performance suffers. 
In general, larger models trained on larger multilingual datasets can improve cross-lingual task performance.
Similarly, scaling the size of the multilingual vocabulary also have a positive effect on cross-lingual performance.
%XLM-R also showed the possibility that multilingual models can be competitive with monolingual models of similar capacity, since it was only slightly surpassed by RoBERTa in GLUE and XNLI benchmarks. 

% diagram of language content comparison from XLM-R
%\subsubsection{ELECTRA}

%``Efficiently Learning an Encoder that Classifies Token Re-placements Accurately'' or ELECTRA \citep{electra_ClarkLLM20} approach used a different pre-training objective compared to BERT.
%BERT's pre-training objective of Masked Language Modeling (MLM) corrupts the input as it replaced some tokens with a [MASK] token then learns to predict the original tokens. 
%In contrast, ELECTRA used Replaced Token Detection (RTD) instead of MLM for pre-training. 
%In RTD, some tokens from the input sequence are replaced by sample tokens from a token distribution which is typically generated from a small generator model. 
%The pre-training objective then is to identify if each token is a real input token or a synthetically generated replacement. 

%During the ELECTRA pre-training process, two neural networks are trained, a generator and a discriminator, both comprising transformer encoders. 
%The generator is trained with a maximum likelihood objective to generate a sequence of tokens rather than being trained adversarially to the discriminator.
%The discriminator network then identifies if each token produced by the generator is a real input or a replaced token.
%The generator network is discarded at the end, and only the discriminator is used for downstream tasks.
%diagram from electra paper

\subsubsection{DeBERTa and mDeBERTa}

%DeBERTa models implement several improvements over BERT, RoBERTa and ELECTRA architectures and comprise leading models in multilingual NLP benchmarks.
%Multiple variants of DeBERTa models will be discussed in this section, including a multilingual version, which is used to create baseline results in this thesis.

%``Decoding-enhanced BERT with Disentangled Attention'' or
DeBERTa \citep{he2021deberta} improves upon BERT and RoBERTa architectures with two techniques.
The first is a disentangled attention mechanism, where each word is represented using two vectors (instead of one) to encode its content and position separately.
The second, is the inclusion of absolute token positions in the decoding layer in addition to relative token positions used by BERT.
Both techniques significantly improved the efficiency of model pre-training and performance of downstream NLU tasks.
It was shown that a monolingual DeBERTa model trained on half the data as RoBERTa large model performs better on a range on NLP tasks.
DeBERTaV3 \citep{he2021debertav3} improved this further with ELECTRA \citep{electra_ClarkLLM20} style pre-training, which replaced Masked Language Model (MLM) training objective with Replace Token Detection (RTD).
%, while introducing gradient-disentangled embedding sharing to avoid tug-of-war dynamics in ELECTRA.
%This pre-training task predicts against all tokens in the input, rather than just the masked token in BERT.  
This approach is efficient especially for smaller models while using less training data and less compute \citep{electra_ClarkLLM20}.
%DeBERTaV3 also eliminated the limitations in ELECTRA-style embedding sharing by using gradient-disentangled embedding sharing.
% In ELECTRA, embeddings are shared between the generator network and the discriminator network, each having MLM and RTD pre-training objectives respectively.
% This pulled the gradient updates to embeddings in different directions creating a tug-of-war dynamics during pre-training.
% DeBERTaV3 prevented this problem by using gradient-disentangled embedding sharing, where gradients for the first network (the generator) are calculated only based on the MLM loss, but not on the RTD loss, and gradients to the discriminator are computed based on both MLM and RTD loss, resulting in better pre-training performance.  
%Due to its combined improvements, DeBERTaV3 surpassed performance over DeBERTa and several previous models.
%For comparison, DeBERTaV3 results on the GLUE benchmark in Table \MDnote{230630}{Which table?}. as produced by 

% Table of DeBERTaV3 performance on GLUE
%\subsubsection{mDeBERTa}

While DeBERTaV3 is a monolingual model trained with English Wikipedia and BookCorpus data, \citet{he2021debertav3} also created its multilingual version as mDeBERTa.
The mDeBERTa model adopts the same dimensions as DeBERTaV3 model but is trained on the same CC100 multilingual dataset as XLM-R.
However, unlike prior XLM models, DeBERTaV3 is not pre-trained on parallel data.
Benefiting from improvements in DeBERTaV3 and cross-lingual transfer, mDeBERTa outperforms the previous state-of-the-art model XLM-R  for all languages on the XNLI benchmark.
% Confirmation on SOTA required
%mDeBERTa is the state-of-the-art cross-lingual language model as of December 2021, and is used as a baseline model in our research.
Table~\ref{tab:mdeberta-xnli-results} reproduces the results from \citet{he2021debertav3} for mDeBERTa versus earlier models on the XNLI benchmark.
In this paper, mDeBERTa is used as the foundation model given its competitive performance despite the small model size.

% Table of mDeBERTa performance on XNLI
\begin{table}[h!]
  \centering
    \tiny
    %l|c|c|c|c|c|c|c|c|c|c|c|c|c|c|c|c
    \begin{tabularx}{\textwidth}{l|X|X|X|X|X|X|X|X|X|X|X|X|X|X|X|X}
      \textbf{Model} & en & fr & es & de & el & bg & ru & ru & ar & vi & th & zh & hi & sw & ur & Avg\\
      \hline
      \multicolumn{17}{l}{Cross-lingual transfer}\\
      \hline
XLM&83.2&76.7&77.7&74.0&72.7&74.1&72.7&68.7&68.6&72.9&68.9&72.5&65.6&58.2&62.4&70.7\\
mT5$_{base}$&84.7&79.1&80.3&77.4&77.1&78.6&77.1&72.8&73.3&74.2&73.2&74.1&70.8&69.4&68.3&75.4\\
XLM-R$_{base}$&85.8&79.7&80.7&78.7&77.5&79.6&78.1&74.2&73.8&76.5&74.6&76.7&72.4&66.5&68.3&76.2\\
mDeBERTa$_{base}$&\textbf{88.2}&\textbf{82.6}&\textbf{84.4}&\textbf{82.7}&\textbf{82.3}&\textbf{82.4}&\textbf{80.8}&\textbf{79.5}&\textbf{78.5}&\textbf{78.1}&\textbf{76.4}&\textbf{79.5}&\textbf{75.9}&\textbf{73.9}&\textbf{72.4}&\textbf{79.8}\\
\hline
\multicolumn{17}{l}{Translate train all}\\
\hline
XLM&84.5&80.1&81.3&79.3&78.6&79.4&77.5&75.2&75.6&78.3&75.7&78.3&72.1&69.2&67.7&76.9\\
mT5$_{base}$&82.0&77.9&79.1&77.7&78.1&78.5&76.5&74.8&74.4&74.5&75.0&76.0&72.2&71.5&70.4&75.9\\
XLM-R$_{base}$&85.4&81.4&82.2&80.3&80.4&81.3&79.7&78.6&77.3&79.7&77.9&80.2&76.1&73.1&73.0&79.1\\
mDeBERTa$_{base}$&\textbf{88.9}&\textbf{84.4}&\textbf{85.3}&\textbf{84.8}&\textbf{84.0}&\textbf{84.5}&\textbf{83.2}&\textbf{82.0}&\textbf{81.6}&\textbf{82.0}&\textbf{79.8}&\textbf{82.6}&\textbf{79.3}&\textbf{77.3}&\textbf{73.6}&\textbf{82.2}\\
      \hline         
    \end{tabularx}
    \caption{mDeBERTa results on XNLI test set under cross-lingual transfer and translate train all settings \citep{he2021debertav3} }
    \label{tab:mdeberta-xnli-results}
\end{table} 

\subsection{Monolingual Models for Myanmar}

\subsubsection{Technical aspects of Myanmar language}

\begin{figure}[h]  
	\begin{center}
      \includegraphics[width=0.3\linewidth]{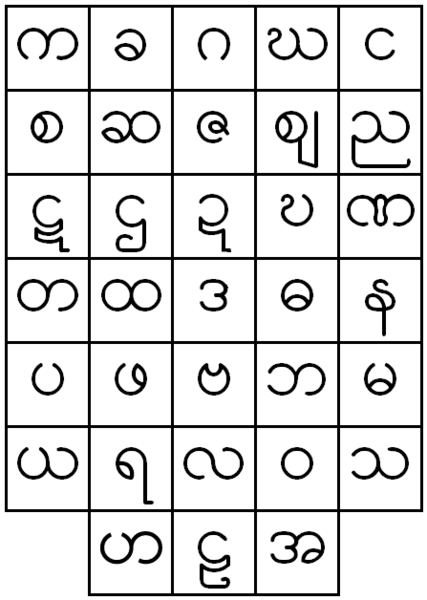}
    \end{center}
	\caption{The 33 consonants of the Burmese alphabet, without diacritics (Wikimedia Commons)}
	\label{fig:burmese-alphabet}
\end{figure}

% background of the language
Burmese or Myanmar language is the national language of Myanmar, a South-East Asian nation with an ethnically diverse population, estimated at over 51 million as of 2019.\footnote{\url{https://www.dop.gov.mm/en/publication-category/2019-inter-censal-survey}}
Although the official English name of the language is Myanmar language, most English speakers continue to refer to the language as Burmese, after Burma, the country's previous and co-official name.
Burmese is the common lingua franca in Myanmar, however it is also spoken in neighbouring regions such as Thailand.
In text form, Burmese is written in Myanmar script, however the Myanmar script is also used for ethnic languages of Myanmar other than Burmese, such as Mon and Shan.\footnote{\url{https://en.wikipedia.org/wiki/Burmese_alphabet}}
It is written from left to right and does not require spaces between words, although spaces are often utilised between clauses to enhance readability and to avoid grammatical ambiguity.
Figure~\ref{fig:burmese-alphabet} shows Burmese alphabets in Myanmar script, without the accompanying modifiers that transform them into words and sentences.

%table of Burmese alphabets
 % https://commons.wikimedia.org/wiki/File:Burmese_alphabet.png

% background of unicode
A key technical challenge with Myanmar script is the non-standardization of fonts and encodings.
Since Myanmar script uses non-Latin alphabets and does not use spaces to separate words, it was not directly supported by ASCII based early systems, and cannot usually leverage Latin based text processing methods.
The Myanmar script was added to the Unicode Standard in Version 3.0 (September, 1999),\footnote{\url{https://www.unicode.org/faq/myanmar.html}} but the implementation of Myanmar script in operating systems, input methods and fonts had lagged behind for several years.
It was not until 2005 that a Unicode-compliant Myanmar font such as \textit{Myanmar1} could be rendered on Microsoft Windows, enabled by the release of Windows XP service pack 2 \citep{battleofthefonts}.
In the mean time, multiple ad-hoc solutions for Myanmar script support have emerged, which are incompatible with the Unicode standard.
One such solution is Zawgyi font and encoding,\footnote{\url{https://en.wikipedia.org/wiki/Zawgyi_font}} which was widely-adopted by the people, catalysed by the major events in Myanmar.
Until recently, Myanmar online users were segregated into two major groups, Myanmar Unicode users and Zawgyi font users \citep{battleofthefonts}.
It has taken large community efforts to unify these groups into using Unicode Standard only, such as the open-source release of Myanmar Tools library\footnote{\url{https://github.com/google/myanmar-tools}} and the introduction of autoconvert feature on Facebook\footnote{\url{https://engineering.fb.com/2019/09/26/android/unicode-font-converter}}.
Some Myanmar language content on the internet can still be found in non-Unicode encoding and may consequently exist in raw datasets created from internet content. 
Since Myanmar language content can exist in two different encodings, Myanmar language solutions also needed to standardise the input before further processing.

% background of burmese NLP
In the field of Myanmar NLP, earlier work used symbolic text processing \citep{word-segmentation-myanmar-2008}, leading up to Statistical Machine Translation methods \citep{myanmar-stats-MT-2016} and explored deep-learning methods towards Neural Machine Translation \citep{myanmar-NMT-2018}  \citep{myanmar-NMT-2019}. 
More recently, the Myanmar NLP community have started to apply monolingual pretrained language models in areas such as POS tagging \citep{myanmarbert_ucsy_2021} and Sentiment Analysis \citep{myanberta_ucsy_2022}.
We explore recent monolingual language models for Myanmar as follows.

\subsubsection{Burmese BERT and ELECTRA}
To our knowledge, the earliest work on monolingual pre-trained models for Burmese was done by \citet{myanmar-bert-electra_jiang_2021}. 
They released four Myanmar specific models based on BERT and ELECTRA architectures.
They used a collection of Myanmar data from the OSCAR corpus,\footnote{\url{https://oscar-project.org}} Common Crawl Corpus\footnote{\url{https://commoncrawl.org}} and Wikipedia to pre-train these models.
Taking into consideration that Myanmar language does not use spaces to separate words, they adopted Sentence-Piece segmentation instead of applying Byte-Pair encoding directly in pre-training.
The models were then fine-tuned and evaluated on POS tagging and text classification tasks.
On both of these tasks, both Burmese BERT and ELECTRA models outperformed previous methods used for Myanmar.

\subsubsection{MyanmarBERT}
\citet{myanmarbert_ucsy_2021} created another BERT model for Myanmar, MyanmarBERT, which is rather trained on a large monolingual corpus in Myanmar.
To pre-train MyanmarBERT, they developed a large Myanmar-only corpus named MyCorpus.
MyanmarBERT was fine-tuned for Part-of-Speech (POS) Tagging and Named Entity Recognition (NER) tasks and compared with Multilingal BERT (mBERT).
Monolingual Myanmar datasets were used to evaluate MyanmarBERT and compared with mBERT.
MyanmarBERT outperformed mBERT on POS tagging but marginally outperformed mBERT on NER.

\subsubsection{MyanBERTa}
Following this work, \citet{myanberta_ucsy_2022} worked on MyanBERTa, another Myanmar pre-trained model based on RoBERTa settings.
MyanBERTa was trained on a large monolingual corpus -- a combination of MyCorpus \citep{myanmarbert_ucsy_2021} and Burmese News and Blog Websites, with over 5M sentences and 136M words.
An additional layer was added to the pre-trained model to create fine-tuned models on NER, POS and Word Segmentation tasks respectively.
It was shown that MyanBERTa outperformed POS and Word Segmentation performance over MyanmarBERT and mBERT, but marginally outperformed on NER.

%Sample body text. Sample body text. Sample body text. Sample body text. %Sample body text. Sample body text. Sample body text. Sample body text.

%\section{This is an example for first level head---section head}\label{sec3}

%\subsection{This is an example for second level head---subsection head}\label{subsec2}

%\subsubsection{This is an example for third level head---subsubsection head}\label{subsubsec2}

%Sample body text. Sample body text. Sample body text. Sample body text. Sample body text. Sample body text. Sample body text. Sample body text. 

\section{Dataset Development Phase 1: Initial Dataset}

\subsection{Overview of myXNLI}

% myXNLI-Dataset-Lineage
\begin{figure}[h]  
	\begin{center}
      \includegraphics[width=0.9\linewidth]{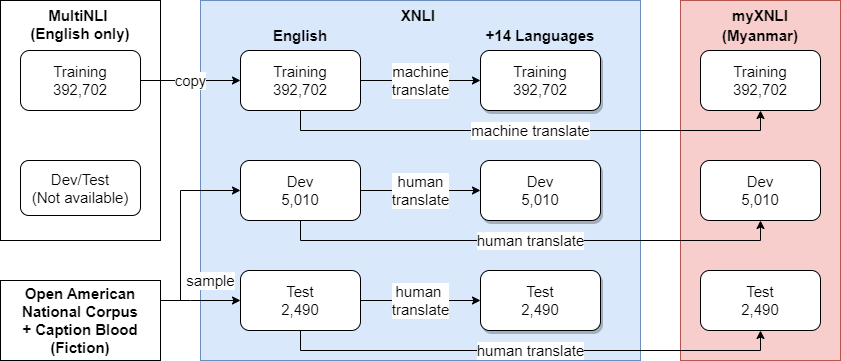}
    \end{center}
	\caption{Lineage between MyXNLI and its parent datasets}
	\label{myXNLI-Dataset-Lineage}
\end{figure}

% Explain the lineage between the datasets
Here we describe the myXNLI dataset, which includes a training set, a validation (dev) set and a test set in the NLI format in Myanmar language.
As shown in Figure~\ref{myXNLI-Dataset-Lineage}, the myXNLI dataset is derived from the MultiNLI \citep{williams-etal-2018-broad} and XNLI \citep{conneau-etal-2018-xnli} datasets.
The MultiNLI dataset provides the original NLI training data in English containing 392,702 sentence pairs, together with the consensus-based labels for each example.
Previous work on XNLI has machine-translated this English training data from MultiNLI to create training datasets in 14 other languages, and reused the English labels directly for the translated data (Section~\ref{ssec:lit-rev-xnli}). 
% \MDnote{230630}{I think you need to explain this a bit more.  See also comments in Ch 2.}
As a natural extension to this, the myXNLI dataset includes the NLI training data in Myanmar which is created by machine-translating the MultiNLI training data from English into Myanmar.
Similar to XNLI, we also reuse the existing labels for English training data for the Myanmar version.

% \MDnote{230701}{In general, you want descriptions of existing work like this to be in past tense. I'm correcting some as I go along, but might miss some.}

As for validation and test set portions, the dev and test sets of the MultiNLI dataset were held private as part of the MultiNLI benchmark evaluation process.
Therefore, the XNLI dataset for dev and test sets used sentences from other English corpora, mainly the Open American National Corpus and Captain Blood (for the Fiction genre).
These sentences in English language were then used to create 7,500 NLI sentence pairs and labeled manually.
To create parallel development and test sets in all XNLI languages, the new English dev/test sets were human-translated into other XNLI languages, and labels from English dev/test sets were also reused for the translated datasets.
We adopted a similar approach for the myXNLI dataset, by translating the XNLI English dev/test sentences into Myanmar. 
For myXNLI, we translated all 7,500 sentence pairs from XNLI English dev/test sets into Myanmar.
The labels from English dev/test sets are also reused for the Myanmar datasets.

\begin{figure}[h]  
	\centering
      \includegraphics[width=0.9\linewidth]{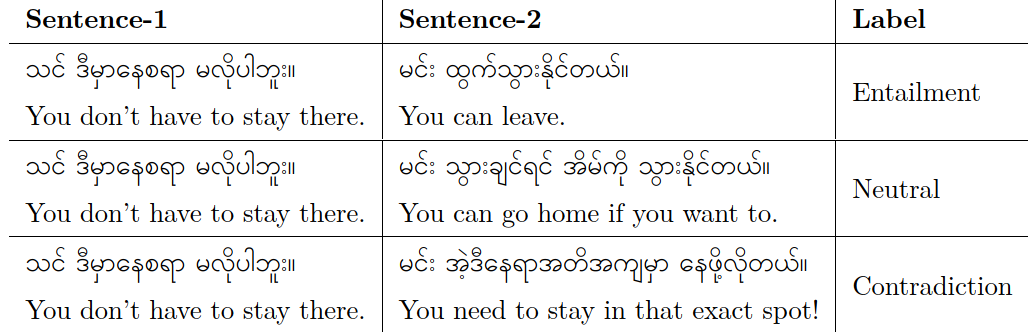}
	\caption{Example myXNLI data in Myanmar and English sentences with labels}
	\label{myXNLI-sentence-pairs}
\end{figure}

Both the machine-translated training set and and human-translated dev/test sets are part of the myXNLI dataset.
Figure~\ref{myXNLI-sentence-pairs} describes an example of myXNLI data in NLI 3-way format.
The label column denotes whether Sentence-1 (Premise) and Sentence-2 (Hypothesis)\footnote{The columns are named Sentence-1 and Sentence-2 in alignment with the XNLI source files, however the sentences are treated as Premise and Hypothesis respectively for the NLI task.} are in an \textit{entailment}, \textit{contradiction} or \textit{neutral} relationship. 
In myXNLI, the English source sentences are kept along side their Myanmar translations.
This is useful for error analysis of NLI models, especially when the Myanmar translations may not explain well why a particular sentence pair is predicted differently than the label, but can be explained from the original English sentences.\footnote{As we will see in Section~\ref{ssec:baseline-discussion}, the quality of translation affects the model outputs.} 
Furthermore, this allowed the dataset to be used for English, Myanmar and cross-matched NLI tasks (Section \ref{par:cross-mapped-nli}).

\subsection{Building the Training Dataset}
% Present or past tenses?

To build the training dataset in Myanmar, we used the English training dataset from MultiNLI containing 392,702 pairs of sentences as our source.
This is also the same dataset from the English portion of the XNLI dataset.
To translate from English to Myanmar, we invoke Google Cloud Translate API\footnote{\url{https://cloud.google.com/translate/docs/reference/api-overview}} from a batch-processing script, with English sentences as the input and the target language set to Myanmar. 
Each MultiNLI example contains two English sentences per line, but some sentences are used in multiple examples (i.e. to make one of entailment, contradiction or neutral sentence pairs each).
Therefore, we translated Sentence 1 and Sentence 2 independently and cache translation results in memory for efficiency.
In the output file for training dataset, Myanmar translations are saved together with the original English sentences, as well as the original labels. 
We then applied light post-processing on the machine-translated output, to clean up invalid tokens such as URL-encoded tokens\footnote{\url{https://en.wikipedia.org/wiki/Percent-encoding}} which would otherwise cause issues in downstream processes.

\subsection{Building the Development and Test Datasets}

% myXNLI-Development-Process
\begin{figure}[h]  
	\begin{center}
      \includegraphics[width=0.9\linewidth]{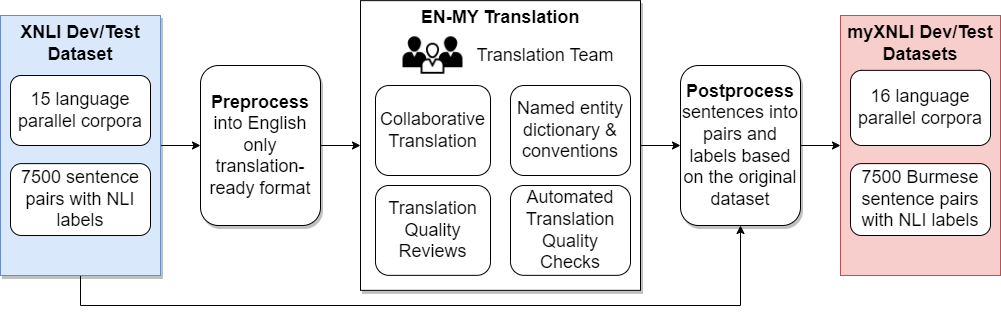}
    \end{center}
	\caption{Workflow for myXNLI Dev/Test Dataset Human Translation}
	\label{myXNLI-Development-Process}
\end{figure}

% \MDnote{230701}{I'm OK with generally using passive voice, but there are some times when it obscures who is doing the thing (which matters when it's you doing something for this thesis).  You'd want to have this section say something like ``We / I / the thesis author / \ldots coordinated the translation effort.  This included both recruitment and management of translators, and development of the environment for managing translations, including a number of scripts and tools to support efficient translation.''  Be clear about your own role, including whether you did any of the translations, etc.}

\paragraph{}
We built the Myanmar development and test sets by carrying out human translation of XNLI English dev/test sets into Myanmar.
Our efforts to build the dataset include the recruitment of translators, setting up a translation environment, defining translation procedures and development of scripts and tools to manage the translations and build output files.
One author also participated in the translation and revision of translations as part of the translation and QA team.
Our end-to-end workflow to build the dev/test sets is described in Figure~\ref{myXNLI-Development-Process}.   

% \MDnote{230701}{You'll want to discuss the recruitment some more.  First, note why you don't use crowdsourcing, since that's the most common thing to do in NLP now --- essentially, lack of necessary expertise.  Then say something about the background of your volunteers --- academics, whatever --- noting any particular issues, perhaps given political situation; you do want to get across what the challenges might be in low-resource situations. Also, maybe list those volunteers who would like to be named in the Acknowledgements section of your thesis, and note that here.}

\paragraph{Translator Recruitment}
Our translation efforts were managed as a project itself as they involve coordination between several translators.
%There were two main phases of translations and corresponding recruitment activities over the course of the project.
In recruiting translators to initiate the translation project, we invited local NLP researchers in Myanmar to collectively translate English data into Myanmar as an open-source project, resulting in the initial version of myXNLI dev/test set.
Working with local NLP researchers under an open-source arrangement has several benefits over other options such as hiring professional translators.
Firstly, Myanmar professional translators are rare, their skills and backgrounds vary, and conventional translators may not be familiar working with file formats and annotations often required in building NLP datasets. 
We also had little translation and annotation guidelines for them to start with at the very beginning, so professional translators' outputs may be sub-optimal.
By inviting local NLP researchers as translators instead, we drew on their prior experience in building former Myanmar-inclusive datasets such as the Asian Language Treebank \citep{alt-thu-etal-2016}, and were able to leverage some previous work, such as general translation guidelines and Myanmar spelling standards.
Secondly, taking the community-based crowd-sourcing approach allowed us to explore how low-resource datasets may be built under limited funding while relying mainly on community contributions.
Last but not least, starting as an open-source project ensures that the NLP community can benefit from the dataset, without any proprietary restrictions imposed.
% \MDnote{230729}{Maybe introduce earlier --- intro of Ch 3? Ch 1? --- that you're open-sourcing the dataset and it has this website.  Currently just noted here, where you're just talking about volunteers, and Acknowledgements.}

\paragraph{Team Profile}
The founding team of myXNLI translators included four local NLP researchers and one of the authors, making a total of five translators.
However this group was eventually extended with eight NLP students to meet the translation workloads.
In this transition, the founding team remained as the core team contributing most of the translations through file submissions and discussions.
The extended team contributed a limited set of translation files assigned to them by the core team.
As a general profile observed by the author, the local NLP group is highly fluent in Myanmar as it is their first language.
They speak Myanmar on everyday situations as well as use it in official and academic contexts in both written and spoken forms.
English, however, is their second language and is used mainly in academic contexts only, and much more in written than spoken forms.
This profile will become relevant to later discussion of the type of translation errors we found in the translation revisions (Section~\ref{ssec:issues}).
% The second phase to recruit expert translators and freelancers to improve the results from the first phase is covered in detail in Section ~\ref{par:expert-translators}.
% Towards the end of the project, the combined translation team included over 20 volunteers (including the author) and 2 freelance translators. 

\paragraph{Creating Translation Files}
% Sourcing into Translation Files 
% Github project etc.
The source XNLI dataset included a parallel corpus with individual sentences used in NLI pairs, for English and other translations.
This contains 10,000 unique English sentences from all 7,500 NLI examples in the XNLI dev/test set. % (Foot note)
We used this corpus as the starting point to create Myanmar translations, rather than directly translating the NLI sentence pairs.
Using scripts, we created 100 translation files from the source corpus, with each translation file containing 100 translation entries.
Once the translation files were initialised with placeholders for Myanmar translations, we uploaded them to a Github repository and made them accessible to the translation team.

\begin{figure}[h]  
	\begin{center}
      \includegraphics[width=0.9\linewidth]{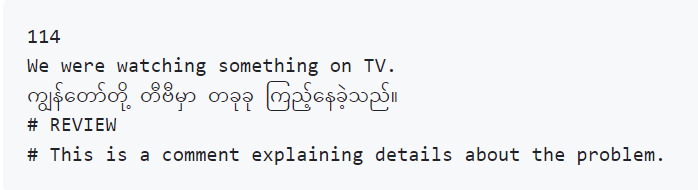}
    \end{center}
	\caption{Example of an entry in myXNLI Translation Files}
	\label{Translation-Entry-1}
\end{figure}

\paragraph{Translation File Format}
An example entry in a translation file is described in Figure~\ref{Translation-Entry-1}.
The first line makes a reference to the line number of the English sentence in the XNLI corpus.
The second line contains the actual English sentence to be translated.
The third line is reserved for Myanmar translation of the English sentence, and initially populated with a placeholder to indicate the human translators.
Additional and optional lines for human translator notes are also allowed with a hash prefix (\#). 
This is useful for flagging translations that require review or documenting any observations made during translation.
Lastly, a blank line is used to separate the current entry from the next.

% Translation Entry 1

% Translator coordinations

\paragraph{Collaborative Translation and QA}
We coordinated the translation efforts using a number of tools and scripts for efficiency whenever possible.
Each translator was assigned a number of translation files to work on and tracked with a shared spreadsheet.
We set up a fortnightly meeting for the core translation team to define translation standards and procedures as well as review challenging translations marked by each translator using the annotated comments.
To maintain the quality of translations across different translators, we established common translation guidelines such as how to translate acronyms, gender pronouns and subject-matter terminologies, without adding additional context to the sentence (Section~\ref{ssec:guidelines}).
The translators also curated a bespoke dictionary to share translations of English phrases and terms which are hard to translate or not easily found in Myanmar literature. 
We developed a validation/QA script to search any remaining entries with missing, incomplete, or inconsistent translations according to the dictionary.
We used Github to version control the translation files, manage access for the translators and track their commits.
A Translation Leaderboard was established to track the progress of translations and individual translators over time.
Once the raw translations were complete, we redistributed the translation files among only the core translation team to revise them.
%\MDnote{230729}{``\ldots core translation team, who revised them''??}
% \MDnote{230701}{Following the comments above, I'd also suggest a bit more detailed data on the translation effort: How many translators? How many translations did each do (bar chart?).}
% Barchat to be added if time allows

% Post-processing to Create pairs and labels
\paragraph{Building the Dataset Files}
When all translations files containing 10,000 entries were completed, they were used to create an English-Myanmar translation dictionary at a sentence level.
Using the original XNLI English dev/test corpus files and this translation dictionary as inputs, we generated output files in NLI format containing Myanmar translations of Sentence-1 and Sentence-2 respectively.  
The original English sentences, NLI labels and Genre labels from the input were also copied across to the output files in this process.
In addition to this dev/test dataset, we also appended Myanmar translations to the XNLI 15-language parallel corpus, to create a 16-language parallel corpus.

\subsection{Translation Guidelines}
\label{ssec:guidelines}

% General translation guidance (ALT)
We established several translation guidelines to keep our translation quality and style consistent throughout the dataset.
Our general guidelines are based upon the Myanmar translation instructions from Asian Language Treebank \citep{alt-thu-etal-2016} and are listed below.

\begin{enumerate}
    \item Don't miss any information in translation.
    \item Don't add unnecessary information.
    \item Take care to minimise spelling mistakes in Myanmar sentence.
    \item Use the spoken or written style of Myanmar, depending on the English sentence. % (e.g. သည်၊ မည်၊ vs. တယ်၊ မယ်)
    \item If possible, use the Myanmar terms that directly align to English (i.e. avoid idiomatic translation).
\end{enumerate}

In addition to the general guidelines, we also developed a number of translation conventions for dealing with specific cases as follows. 

\begin{itemize}
    \item \textbf{Past tenses and Gender Pronouns} are often optional in Myanmar and are mainly used in formal writing style. 
We include modifiers for past tenses and gender specific pronouns when the source sentence is in formal writing style, and omit them when the sentence is in spoken style.
    \item \textbf{Exclamations and Interjections} are translated to equivalent Myanmar terms whenever possible, otherwise transliterated into Myanmar.
    \item \textbf{Acronyms, Measurement Units and Symbols} are translated into equivalent Myanmar words when they exist, otherwise English notations have been reused.
    \item \textbf{Named Entities and Scientific or Subject Specific Terms} are translated into Myanmar if there are adaptations in Myanmar, but otherwise transliterated into Myanmar (initial rule). But we found out eventually that such transliterations introduced inconsistencies in the output (Section \ref{ssec:issues}), so as a final rule, we kept these phrases as English if an appropriate Myanmar adaptation is not found.
    \item \textbf{Quoted text} such as Latin or French in the source sentence are repeated as-is in the translated sentence.
\end{itemize}

\subsection{Additional Machine Translated Test Sets}
\label{ssec:translate-test}

% \MDnote{230701}{Note the XNLI paper's comparable setup, either here or earlier in Ch 2 (and then refer back here).}

Previous XNLI results in \citet{conneau-etal-2018-xnli} and \citet{conneau-etal-2020-unsupervised} included a
translate-test evaluation scenario, where non-English test data is translated into English and an English-only model is used. 
This scenario can provide an evaluation baseline on a model's performance in a low-resource language, compared to using machine translation into a high-resource language first then performing the task using a high-resource monolingual model.
To align and compare myXNLI results with XNLI results, we also generated machine-translations of myXNLI test set from Myanmar back into English.\footnote{Although we acquired the Myanmar test set by human-translation from English.} 
While not part of myXNLI dataset itself, the translate-test dataset helped us evaluate the effectiveness of multilingual models fine-tuned on myXNLI and hence the usefulness of the dataset itself.

% \MDnote{230701}{Say something here in line with a discussion we had during one meeting: I think I looked up the size of Wikipedias in each language, and Burmese ranked somewhere between these two; use this to illustrate what it means to be low-resource.} 

While our main focus was to evaluate model performance in Myanmar language, we also planned to compare Myanmar results with other low-resource languages.
For this comparison, we selected Swahili (sw) and Urdu (ur) as the two reference low-resource languages.
We selected these languages because they are already part of XNLI languages with matching datasets to myXNLI, and also they have comparable low-resource statistics to Myanmar.
Using Wikipedia as a representative example, Myanmar, Swahili and Urdu each have number of Wikipedia articles between 50,000-200,000, while high-resource languages such as English have articles over 6 million.
In fact, Myanmar is between Swahili and Urdu in terms of ranking by Wikipedia size (number of articles) as described in Table~\ref{tab:wikipedia-sizes}.\footnote{\url{https://meta.wikimedia.org/wiki/List_of_Wikipedias}}
Swahili and Urdu are thus comparable in terms of levels of resources with Myanmar, but differ in other potentially interesting respects: for instance, Swahili is written in Latin script, while Urdu, although written in its own script (derived from Persian), has a much more standardised romanization.

\begin{table}[h!]
%  \begin{center}
    \begin{tabular}{l|c|c}
      \textbf{Rank} & \textbf{Language} & \textbf{Number of Articles}\\
      \hline
      1 & English & 6,685,265\\
      \hline
%      2 & Cebuano & 6,122,993 \\
%      \hline
      3 & German & 2,818,210\\
      \hline
      5 & French & 2,557,357\\
      \hline      
      55 & Urdu & 192,793\\
      \hline 
      71 & Myanmar & 106,770\\
      \hline 
      83 & Swahili & 78,307\\
      \hline         
    \end{tabular}
    \caption{Wikipedia sizes between some high-resource and low-resource languages (July, 2023)}
    \label{tab:wikipedia-sizes}
%  \end{center}
\end{table} 

To create the translate-test datasets in Myanmar, Swahili and Urdu, we followed a similar machine translation process used earlier for the Myanmar training set, i.e. by running a batch-script invoking Google Cloud Translate API to translate the test sets of target languages back into English.
In Section~\ref{ssec:baseline-discussion}, we provide these evaluations on the translate-test datasets using monolingual models.

\section{Dataset Development Phase 2: Revised Dataset}

\subsection{Common Issues from the Initial Version}
\label{ssec:issues}

The initial version of the Myanmar dev/test set included several mistranslations.
Many of these errors were not identified during the first round of translation QA, where the same group of translators exchanged translation files between themselves and revised them.
This led us to another round of translation QA, where we engaged a group of individuals with higher bilingual skills to correct and rate the translations.
From this expert review, we have categorised and discussed common translation errors as follows.

% \MDnote{230701}{You might want to say a bit more about the versioning process.}

\begin{itemize}
    \item \textbf{Mistranslations of polysemous English words:}
For example, the word \textit{reach} in English can be used to describe reaching to a destination, as well as reaching out to a person.
The same can be said about the words \textit{right} and \textit{right now}.
Based on the actual context in the sentence, such phrases must be often translated into different terms in Myanmar.
However, an inexperienced translator  may hastily assume the more common meaning of the word and mistranslate in the target sentence.
% \MDnote{230701}{This is why it will be useful earlier to give a bit of background on your translators.  How much translation experience do they have, and to what extent are they bilingual? This is something that developers of resources for low-resource languages will have to deal with, so it will be useful to document.}
    \item \textbf{Arbitrary transliterations of English named entities:}
We found that arbitrary transliteration of English names into Myanmar can lead to inconsistent spellings across the corpus, and can lead to unexpected results when both inconsistent names appear in a NLI sentence pair.
Therefore we urged translators to reuse English named entities as-is in Myanmar sentences, unless they already exist in Myanmar literature. 
For example, we required that the name \textit{England} should be transliterated into Myanmar as there is already a standard Myanmar spelling for this.
But infrequent named entities like \textit{James Whitcomb Riley}, \textit{Eugene V. Debs} and \textit{Madam C.J} should be just kept as English in the Myanmar translation.
     \item \textbf{Inadequate cultural or background knowledge:}. 
For example, when given the word \textit{Indians} most Myanmar translators will translate this as South Asian Indians, since India is the neighbouring country of Myanmar. However, the original sentence may very well refer to Native Americans, which is a concept much less frequently used in Myanmar language.
In another example, \textit{Scotches on the Rocks} and was translated word-to-word in Myanmar, since the translator was not familiar with how whiskey is consumed in the western culture. 
An idiom \textit{The whole nine yards} was translated directly to Myanmar for a similar reason. 
Mistranslations due to lack of cultural or background awareness are more challenging to spot and required revision by experienced translators.
    
\end{itemize}

\subsection{Expert Translation Revision}

% \MDnote{230701}{Again, a bit more data here.  How many expert translators, and how many did each review?}

\paragraph{Expert Translators}
\label{par:expert-translators}
As mentioned earlier, we observed that many of the mistranslations are mainly due to the variation in bilingual proficiency between translators.
To rectify this, we added an expert translation revision phase for the dev/test set, where we only invited proficient translators to review and correct existing translations.
This group of expert translators included two freelance translators as well as eight bilingual volunteers (including one of the authors), who speak Myanmar at home but use English professionally.
All volunteers have similar backgrounds: they were born and raised in Myanmar and speak Myanmar at home, but they also studied a University degree and/or work professionally in English-speaking countries.
We believe this bilingual background is an important trait to recognise the mistranslations and improper usage of Myanmar words.
Each volunteer was given a batch of five files,\footnote{Some keen volunteers contributed up to six files.} with many of them requiring at least one week to complete each file.\footnote{They volunteered in their personal time after work.}
We also recruited two freelancers as the number of volunteers did not cover the entire volume of translation files.
Freelancer 1 worked on 36 files and Freelancer 2 worked on 15 files.
We found that it was a good idea to work with more than one freelancer in parallel, as this allowed us to balance their translation quotas based on changing circumstances.
 
%A requirement for all expert translators is to be tech-savvy enough so that they can use translation and coordination tools such as Github, Google Sheets and Google Meet.

\paragraph{Revision Process}
We archived the earlier version of the dataset before the expert revision on the existing Github repository as a branch, but also created a new Github repository to exclusively on-board the expert translation team and kept their git commits separate from the original translation team. % how about named folders?
On this separate Github repository, we provided a comprehensive README file with examples of translation errors and how to correct them.  
We believe having clear instructions from the beginning becomes more important in this phase because it would be challenging to setup online coaching meetings with translators living across multiple English-speaking countries and different time zones.
This is in contrast to previous translation phase, where all translators live in Myanmar and group discussions are necessary to establish translation guidelines.
The expert translators independently reviewed each translation file to correct or improve Myanmar translations whenever possible.
Another rectification in the revision phase was to consistently reuse English named entities as-is in Myanmar sentences, unless they already exist in Myanmar literature. 
% Moved to a new repo
% Given a readme file
% Tracked by Spreadsheets
% Weekly check-in

\paragraph{Translation Ratings}
The expert translators also rated each final translation using a 1-5 point Likert scale, with a rating of 1 as the lowest quality and as 5 being the perfect translation. 
These ratings were given by each translator on their own work to capture their confidence in each translation and annotate challenging sentences.
%\MDnote{230729}{Did each expert translator rate his/her own translation or another expert translator's?}
It is also possible to improve the accuracy of these ratings further by cross-checking and rating each other's work.
Having a rating for each translation gives us a sense of overall translation quality of the dataset. 
Table~\ref{tab:translation-rating} describes the rating system used for the final translations as well as the distribution of translation quality across the dataset after the expert revision.
For many translations, the expert translators were able to improve their quality and correct the translation mistakes from the prior group.
However, the quality of each translation is also directly related to the quality of the original English sentence.
Some English sentences were too hard to translate precisely even for the expert translators to get perfect translations.
For a few English sentences which are not complete sentences or provide sufficient context, the corresponding Myanmar translations are also incomprehensible(0.8\% or 80 out of 10k sentences).

% {5: 5168, 4: 3950, 2: 197, 3: 606, 1: 80}
\begin{table}[h!]
\centering
\footnotesize
\begin{tabular}{p{0.1\linewidth} | p{0.7\linewidth} | p{0.1\linewidth}}
    \textbf{Rating} & \textbf{Description} & \textbf{Distribution}\\
    \hline
    5 &  Translation is perfect. & 51.68\%\\
    \hline
    4 & Translation is somewhat unnatural, but the overall meaning is correct. &  39.50\%\\
    \hline
    3 &  Translation is partially correct, but missing some details. &  6.06\%\\
    \hline
    2 &  Translation is wrong or misleading &  1.97\%\\
    \hline
    1 &  Translation is incomprehensible. & 0.80\%\\   
    \end{tabular}
    \caption{Ratings given to myXNLI dev/test set translations during final revision phase}
    \label{tab:translation-rating}
\end{table} 

% Myanmar XNLI Dataset standardization.  Does it improve performance before and after standardization?

\subsection{Identifying Semantic Changes After Translation}
%\subsection{Identifying the Effects of Translation}

% \MDnote{230701}{Explain why they did the English ones --- to establish a baseline level of agreement that might be expected in re-annotating labels with no language transfer issues.}

Similar to XNLI, myXNLI reused the labels from the original English dataset for the translated Myanmar dataset.
\citet{conneau-etal-2018-xnli} studied whether the semantics of NLI sentences in the target corpus could change occasionally as a result of information added or removed in the translation process, and would imply a different label than the original.
To confirm this, they recruited two bilingual annotators to relabel 100 English and French examples each and compared with the original labels.
English examples were relabeled in this case to establish a baseline level of agreement between the new annotator and the original without introducing language transfer issues.
They found that almost all of the labels and semantic relationships between the two languages have been preserved in spite of the translation.
Following a similar approach, we also recruited two bilingual annotators initially to relabel 100 examples each in both English and Myanmar.
We drew these samples from random subsets in the development set. 
After finding that our initial reconciled numbers are lower than \citet{conneau-etal-2018-xnli}, we recruited two more annotators to relabel the same sample sets and received similar results.
On average, the new labels matched the original gold labels 71.25\% for English and 63.25\% for Myanmar, with a gap of 8\%.
These numbers are a bit lower than \citet{conneau-etal-2018-xnli}, and the gap is a bit larger, but still provide some confidence in the labelling while indicating that there is a small amount of label change from the translation.
%\MDnote{230729}{These numbers are the 77\% and 70\%, right?  If so, comment a bit on these.  E.g. 
% what are the examples that changed?

\begin{table}[h!]
\footnotesize
\label{tab:translation-relabelling}
\begin{tabular}{c|c|c|c|c|c}

    \textbf{Lang} & \textbf{Person-1(Set-1)} & \textbf{Person-2(Set-1)} & \textbf{Person-3(Set-2)} & \textbf{Person-4(Set-2)} & \textbf{Avg}\\
    \hline
    English & 67\% & 69\%   & 72\% & 77\% & 71.25\% \\
    \hline
    Myanmar & 59\%  & 59\%  & 65\% & 70\% & 63.25\%\\
    \hline
\end{tabular}
    \caption{Reconciliation (matches) between gold labels and new labels on sample pairs}
\end{table} 

% In a similar experiment, we also investigated the examples of on pronouns dropped during translation to Myanmar.
%e. Amanaki et al. (2022) noted that during translation of XNLI from English to Greek version, many pronouns were dropped due to the
%pro-drop nature of the Greek language. 
%XNLI corpus by adding the dropped pronouns back to compare differences between
%the two Greek versions of the corpus. Subsequently it was found that NLI performance
%dropped significantly when the premise was used from the original version and hypoth-
%esis was used from the de-dropped version, and vice versa. 

\subsection{Quality of Machine Translation}

Since the training portion of myXNLI is machine-translated, the quality of machine translation can significantly impact the quality of the training data. 
To assess the machine translation quality of the training dataset, we used the BLEU score of the same machine translation system over the test dataset as a proxy indicator.
The BLEU score \citep{papineni2002bleu} is commonly used to indicate the quality of machine translation outputs, and is calculated by comparing the candidate translations to to the reference translations.
The human translations for the dev/test dataset are already available from the dataset development earlier, hence can be used as reference translations in this process.
In particular, the test dataset contains 5,000 NLI sentence pairs containing 6,679 unique sentences in both English and Myanmar parallel representations. 
To get the candidate translations, we used the same Google Cloud Translate API to get alternative Myanmar translations of these English sentences.
An additional step required to compare machine-translated and human-translated Myanmar sentences is to break down each sentence into meaningful tokens for comparison.
Unlike English, Myanmar script does not use spaces to separate between words, thus an algorithm is necessary to break Myanmar phases into syllables or words.
For this reason, we used an opensource library Sylbreak\footnote{\url{https://github.com/ye-kyaw-thu/sylbreak}} to get Myanmar syllables for each sentence. 
After converting into syllables, the machine-translated sentences (candidates) are compared to human-translated sentences (references). 
Over the entire test dataset, we obtained a corpus level BLEU score of 51.73.
Since the genres in the training dataset are similar to genres in the test dataset (with test dataset containing more genre variations), we consider this score to be a good indicator of machine translation quality in the training dataset.
For comparison, an earlier English to Myanmar Neural Machine Translation system by \citet{wang-etal-2019-english} obtained a BLEU score of 19.73 on a different corpus and evaluation dataset.
For our case, we believe the BLEU score is significantly higher due to myXNLI dataset containing relatively simpler and shorter sentences for NLI purposes, combined with with significant improvements in machine translation methods over the recent years.
Additionally, to compare this with Google Translate's quality for other low-resource languages over the same corpus, we evaluated the BLEU scores for English to Swahili and Urdu translations using the XNLI test data, and obtained the BLEU scores of 23.73 (English to Swahili) and 29.05 (English to Urdu).
This suggests that the machine translation quality for Myanmar in myXNLI training data is competitive for a low-resource language.

\section{Initial Baselines on Myanmar XNLI}
\label{sec:baselines}

\subsection{Evaluation Approach}

The resulting myXNLI dataset allowed us to train and evaluate language models on Myanmar and English NLI tasks.
Along with the dataset, we provide evaluation results on a selection of language models under a number of scenarios comparable to previous XNLI benchmarks. 

\paragraph{Model Selection}
The language models we selected for the baseline evaluation include XLM-R, mDEBERTa and their monolingual variants. 
Both model architectures had previous XNLI scores to compare to, allowing us to have references in other languages.
XLM-R was chosen since it was one of the first models discussed for cross-lingual-transfer at scale with established XNLI scores \citep{conneau-etal-2020-unsupervised}.
Additionally, we chose mDeBERTa as a successor to XLM-R, which outperformed other models and became state-of-the-art at the time \citep{he2021debertav3}.
For English monolingual models, we chose RoBERTa and DeBERTaV3 as monolinugal counterparts for XLM-R and mDeBERTa respectively.
For Myanmar monolingual model, we chose MyanBERTa as Myanmar monolingual counterpart for XLM-R.

\paragraph{Language Selection}
Our experiments cover four languages, English (en), Myanmar (my), Swahili (sw) and Urdu (ur).
We include English as default high resource language, where the training data and monolingual results for comparison were already available.
Myanmar is the primary language that myXNLI offers where no prior benchmarks exist, and is therefore the focus of our experiments.
We also include Swahili and Urdu as references in other low-resource languages, with previously established XNLI benchmarks (Section~\ref{ssec:translate-test}).
We used XNLI for English, Swahili and Urdu test sets. 

\paragraph{}
Using the models and languages selected, we ran our experiments under the following scenarios to be comparable to previous XNLI benchmarks in \citet{conneau-etal-2018-xnli}, \citet{conneau-etal-2020-unsupervised} and \citet{he2021debertav3}. In addition, we added a scenario to compare monolingual model vs. multilingual model performance for Myanmar.
% \MDnote{230701}{Again, relate these scenarios to what they did in the XNLI paper.}

\paragraph{Cross-Lingual Transfer} 
In this scenario, we evaluate how a model only fine-tuned for high-resource language can perform on low-resource languages via cross-lingual transfer.
For this, we fine-tuned XLM-R and mDeBERTa on English and evaluate on the test sets in myXNLI for Myanmar and in XNLI for Swahili and Urdu.
%\MDnote{230729}{Maybe ``evaluate on the test sets in myXNLI for Myanmar and in XNLI for Swahili and Urdu''.}

\paragraph{Translate Test} 
In this scenario, we evaluate how a model performs directly on a low-resource language compared to using machine translation first into a high-resource language, then solve the task with a high-resource monolingual model.
This could also be considered as a fallback option when a model for a target language is not available.
We used English monolingual models RoBERTa and DeBERTa as they have comparable architectures to XLM-R and mDeBERTa. 
We fine-tune these monolingual models on English only and evaluate them using English data only.

\paragraph{Translate Train}
In this scenario, we evaluate how well a multilingual model can be fine-tuned for Myanmar.
We fine-tune XLM-R and mDeBERTa on Myanmar using myXNLI training set, then evaluate Myanmar NLI performance using the test set.

\paragraph{Monolingual}
In this scenario, we also evaluate how well a monolingual model in a low-resource language can be fine-tuned for the NLI task in the same language.
We consider MyanBERTa for this experiment, as it has a comparable architecture to RoBERTa and XLM-R. 
To evaluate monolingual model performance, we fine-tune MyanmarBERTa on myXNLI training set, then evaluate on the corresponding test set.

\paragraph{Evaluation Parameters}
In our experiments, we used base size models for all model architectures.
In alignment with the XNLI benchmark, we used \textit{accuracy} as the metric for our evaluation.
For all languages and scenarios, the corresponding training sets (392,702 examples each) were used to fine-tune the models, while validation sets (2,500 examples each) were used during fine-tuning as a guide to monitor true performance. 
All experiments involve fine-tuning the models for a single epoch only. 
We found that fine-tuning for subsequent epochs did not improve the accuracy further.  
The performance scores were obtained on the test sets (5,000 examples each) in each language.
Although XNLI results for English, Swahili and Urdu are already available in the community for XLM-R and mDeBERta, we repeat the experiments for those languages under similar settings and hyper-parameters as Myanmar evaluations for consistency.

% \MDnote{230701}{Did you mean devsets here?  That's what you should be monitoring.}

\subsection{Baseline Results and Discussion}
\label{ssec:baseline-discussion}

% \MDnote{230701}{This paragraph should go under Evaluation Approach.}

In this section, we present baseline myXNLI/XNLI results for our experiments over English, Myanmar, Swahili and Urdu languages.
For Myanmar, these are the very first NLI benchmark results enabled by the myXNLI dataset.
Our baseline results are shown in Table~\ref{tab:baseline-myxnli} and discussed below.
%\MDnote{230730}{I'd put best result in table in bold.}

\paragraph{For Initial and Revised myXNLI versions}
For Myanmar, we include two scores side-by-side to differentiate the results from the two versions of the dataset under the same scenario.
Results form the initial myXNLI version is described in parentheses, while results from the revised and final myXNLI version is described without the parentheses. 
We found that the revised version of the myXNLI dataset provides better Myanmar results under every evaluation scenario, by up to 2 percentage points of accuracy.
There is a clear association between the translation quality of myXNLI dataset and the NLI performance scores for Myanmar.

\paragraph{Cross-lingual Transfer}
We found that the XLM-R and mDeBERTa fine-tuned on English XNLI data can provide reasonable results on Myanmar without further fine-tuning.
Cross-lingual transfer performance on Myanmar is relatively higher than Swahili and Urdu. 
For English, Swahili and Urdu, the scores we obtained are lower than previous results from \citet{conneau-etal-2020-unsupervised} and \citet{he2021debertav3}.
In general, models in their work achieved slightly better scores (2-3 points higher) than models in our experiments.
We also noted that between Swahili and Urdu, the score for Urdu is higher for XLM-R but lower for mDeBERTa (but for Swahili, the opposite).
This observation is also consistent with the the results from \citet{conneau-etal-2020-unsupervised} and \citet{he2021debertav3} and highlighted in Table~\ref{tab:cross-lingual-comparison}.

\paragraph{Translate Test} 
Using machine-translated test sets, we evaluated the selected monolingual models RoBERTa and DeBERTaV3 for English.
Our translate-test scores for Myanmar, Swahili and Urdu are higher than cross-lingual transfer scores. 
%\MDnote{230729}{Why aren't there a phase 1 vs phase 2 results for Myanmar here? You sill have to translate a version of myXNLI to English to do this, right? If I'm wrong, you'll still want to explain why the numbers aren't there.}
This suggests that using a combination of machine translation and high-resource monolingual models could be a reasonable alternative in some low-resource situations.
In comparison with previous work, RoBERTa translate-test scores for Swahili and Urdu from \citet{conneau-etal-2020-unsupervised} are lower than our translate-test scores, as well as their own XLM-R cross-lingual transfer results.
We suspect that this is due to an older machine-translation system they used to acquire their translate-test datasets at the time of their experiments, compared to the much more recent machine-translation tool that we used (Google Cloud Translate API).
For Myanmar, we provide two translate-test scores corresponding to the English translations before and after the revision of the Myanmar dataset.
We achieved better translate-test accuracy after the revision of Myanmar dataset, suggesting that improving Myanmar translations in the testset improved the machine-translated English outputs and in turn lifted the English model performance. 
% We do however note that the two versions of translate-test are done a few months apart (before and after the dataset revision) 
% \MDnote{230701}{Can you compare this to the source papers or the XNLI paper?}

% Note this is different from translate train all
\paragraph{Translate Train}
For this scenario, we fine-tuned XLM-R and mDeBERTa models on myXNLI and evaluated on the same.
We found that the best scores in Myanmar NLI performance across all scenarios are achieved by fine-tuning mDeBERTa on Myanmar.
While there is no previous NLI result for Myanmar to compare, we refer to previous translate-train-all results for XLM-R and mDeBERTa from 
\citet{conneau-etal-2020-unsupervised} and \citet{he2021debertav3} respectively, where they fine-tuned the models on all 15 XNLI datsets.
In their Swahili and Urdu XNLI results for XLM-R and mDeBERTa respectively, translate-train-all approach out-performs cross-lingual-transfer and translate-test approaches, and this is consistent with our translate-train results for Myanmar, although we fine-tuned our model for a single language (Myanmar) only.
It is also consistent with the superior performance of translate-train in \citet{he2021debertav3} over cross-lingual transfer (as shown in Table~\ref{tab:mdeberta-xnli-results}).
For RoBERTa and XLM-R however, translate-test on the English-only model (RoBERTa) still outperforms translate-train on the multilingual model (XLM-R) for Myanmar, Swahili and Urdu.
% As mentioned before, we suspect this could be due to the nature of machine-translation system used for translate-test datasets, but also a learning limitation with XLM-R model which we do not observe in mDeBERTa. 
% This suggests that fine-tuning a multilingual model on in-language data is usually better if such a dataset is available.
% This is consistent 
% \MDnote{230701}{Also relate this to other papers.  Was this the best approach in those?}

\paragraph{Monolingual} 
The Myanmar monolingual model MyanBERTa did not perform as well as other models and approaches under our settings.
Further fine-tuning on Myanmar also did not improve the results.
Since monolingual models for Myanmar are still emerging, we expect future models will provide more comparable results.

\paragraph{Overall}
In our baselines overall, fine-tuning mDeBERTa (translate-train) gives the best performance for Myanmar and Urdu, while DeBERTaV3 performs best for English and Swahili (as translate-test).  We note that high-resource English has the highest results for each configuration, as expected.  Of the three low-resource languages, Myanmar has the highest scores, often by some margin, with the other two being broadly similar.  This gives some confidence regarding the quality of the dataset.

\begin{table}[h!]
%  \begin{center}
    \begin{tabular}{l|c|c|c|c|c} % <-- Changed to S here.
      \textbf{Model} & \textbf{Training Data} & \textbf{English} & \textbf{Myanmar} & \textbf{Swahili} & \textbf{Urdu}\\
      \hline
      \multicolumn{6}{c}{\footnotesize CROSS-LINGUAL TRANSFER --- Fine-tune multilingual model on English data}\\
      \hline
      XLM-R$_{\text{base}}$ & English XNLI & 83.2 & 68.42 (68.1) & 64.21 & 66.04\\
      \hline
      mDeBERTa$_{\text{base}}$ & English XNLI & 87.24 & 75.72 (73.89) & 71.77 & 70.01\\
      \hline
        \multicolumn{6}{c}{\footnotesize TRANSLATE-TEST --- Translate everything to English and use English-only model}\\
        \hline
        RoBERTa$_{\text{base}}$ & English XNLI & 85.5 & 76.94 (74.69) & 74.15 & 70.49\\
        \hline
        DeBERTaV3$_{\text{base}}$ & English XNLI & \textbf{90.5} & 78.36 (75.78) & \textbf{75.62} & 72.05\\
        \hline         

      \multicolumn{6}{c}{\footnotesize TRANSLATE-TRAIN --- Fine-tune multilingual model on Myanmar data}\\
      \hline  
      XLM-R$_{\text{base}}$ & myXNLI & 79.28 & 74.13 (72.0) & 64.83 & 68.38\\
      \hline
      mDeBERTa$_{\text{base}}$ & myXNLI & 85.34 & \textbf{79.46 (78.06)} & 73.15 & \textbf{73.09}\\
        \hline 
        
    \multicolumn{6}{c}{\footnotesize MONOLINGUAL --- Fine-tune Myanmar-only model on Myanmar data} \\
    \hline
      MyanBERTa & myXNLI & - & 57.40 (55.6) & - & -
    \end{tabular}
    \caption{Baseline NLI evaluation results (accuracy) using myXNLI/XNLI}
    \label{tab:baseline-myxnli}
%  \end{center}
\end{table}

\begin{table}[h!]
%  \begin{center}
    \begin{tabular}{l|c|c|c|c} % <-- Changed to S here.
      \textbf{Model} & \textbf{Training Data} & \textbf{English} & \textbf{Swahili} & \textbf{Urdu}\\
      \hline
      \multicolumn{5}{c}{\footnotesize CROSS-LINGUAL TRANSFER (Previous Work)}\\
      \hline
      XLM-R$_{\text{base}}$ & English XNLI & 85.8 & 66.5 & 68.3\\
      \hline
      mDeBERTa$_{\text{base}}$ & English XNLI & \textbf{88.2} & \textbf{73.9} & \textbf{72.4}\\
      \hline
      \multicolumn{5}{c}{\footnotesize CROSS-LINGUAL TRANSFER (Our Experiments)}\\
      \hline
      XLM-R$_{\text{base}}$ & English XNLI & 82.2 & 64.2 & 66.0\\
      \hline
      mDeBERTa$_{\text{base}}$ & English XNLI & \textbf{87.24} & \textbf{71.77} & \textbf{70.01}\\
      \hline         
    \end{tabular}
    \caption{Comparison of English, Swahili \& Urdu XNLI scores between \citet{conneau-etal-2020-unsupervised}, \citet{he2021debertav3} and our experiments}
    \label{tab:cross-lingual-comparison}    
%  \end{center}
\end{table} 

\section{Improved Results on Myanmar XNLI}

\subsection{Methods for improving XNLI performance}

Continuing from our baseline results, we applied a number of data augmentation methods to improve the Myanmar NLI performance.
We focused our methods on improving mDeBERTa as it is the best performing model.
Aligning with the structure of the baselines, we apply our methods on mDeBERTa and evaluate the results for English, Myanmar, Swahili and Urdu languages.
We start with the cross-lingual model baselines, but in addition to the fine-tuning on English from Table~\ref{tab:baseline-myxnli}, we also fine-tune on the other languages.

\paragraph{Adversarial NLI data augmentation} 

Previous work in \citet{nie-etal-2020-adversarial} showed that English NLI can be improved by training additionally with Adversial NLI data.
Moreover, our baseline results showed that NLI concepts learned in English are transferred to Myanmar and other languages.
Therefore we conjectured that additional fine-tuning on English datasets such as ANLI can indirectly improve on Myanmar.
However, given that neural network models have well-known capacity issues like catastrophic forgetting \citep{kirkpatrick-etal:2017:PNAS}, and per-language capacity suffers as more languages are added to a fixed-size model \citep{conneau-etal-2020-unsupervised}, there may be a negative effect on other languages if additional English training is given.
To evaluate the effect on ANLI training, we fine-tuned mDeBERTa on English XNLI and English ANLI and observe the performance on Myanmar and other languages.
% \MDnote{230730}{I don't think this is valid to say. Instead, something like: `` \ldots''.}, 
% Additionally,  Add Myanmar + Myanmar ANLI results

\paragraph{Multilingual NLI data augmentation} 
\citet{conneau-etal-2020-unsupervised} showed that training on more languages generally improved the NLI performance but occasionally  this results in degraded performance for some languages due to capacity dilution. 
To specifically observe the effects of adding languages one at a time to a model, we designed experiments in fine-tuning on English combined with each of Myanmar, Swahili and Urdu XNLI datasets.
We expect that Myanmar NLI results will be improved by training together with a high-resource language data such as English XNLI.
We also expect to see similar results for Swahili and Urdu under this scenario.
% What are the relationships between which language learned more and which learned less?

\paragraph{Cross-matched NLI data augmentation}
\label{par:cross-mapped-nli}
As an extension of multilingual NLI data augmentation, we also explored data augmentation with mixed language NLI pairs.
We took our inspiration from a community model on HuggingFace which was based on XLM-R and fine-tuned on shuffled XNLI data.\footnote{\url{https://huggingface.co/joeddav/xlm-roberta-large-xnli}}
Keeping original English sentences along with Myanmar translations in the myXNLI dataset enabled us to create NLI pair combinations with either English or Myanmar in premise and hypothesis positions.
Our experiment therefore used quadruple training data, each with sentence pairs in \textit{en-en}, \textit{en-my}, \textit{my-en} and \textit{my-my} combinations.
We fine-tune mDeBERTa on this combined myXNLI dataset and report NLI performance on each language.

\paragraph{Genre as Side-Input}
Previous work by \citet{genre_muller-eberstein-etal-2021} demonstrated using genre to improve dependency parsing. 
In another study, \citet{side-info_hoang-etal-2018} showed that side-information can be used to improve results in machine translation.
For our purposes in Myanmar NLI, we explored if genre labels can be also leveraged as side input.
We suspect that there may be certain characteristics of genre which allow the model to learn slightly better parameters across different genres.
Specifically, our method explored if the NLI task for a given sentence pair can be done better if the genre is first recognised.
Genre metadata is already available in myXNLI for each sentence pair, as it can be sourced from the original MNLI and XNLI datasets.

To design fine-tuning approaches that utilise genre metadata as side input, we took inspiration from \citet{side-info_hoang-etal-2018} where a prefix was added to the input denoting the metadata. 
In our case, the genre label was added as a prefix in the input sentences.
In myXNLI, both sentences in an NLI pair has the same genre.
In mDeBERTa and BERT architectures, these two sentences are encoded separately by the use of sentence masks. 
Therefore, we added the same prefix to both sentences.
Additionally, we suspected that the genre names might occasionally overlap with words in the actual sentences and this may make the training less effective.
For example, the genre label \textit{Travel} could overlap with the actual content of the sentences.
To prevent this situation, we created special tokens for each genre type and gave them distinct embedding values in the mDeBERTa tokenizer.
This process is similar to using dedicated embeddings for special tokens CLS and SEP tokens used by BERT.
Adding genre labels as special tokens also treats them as categorical data, rather than a string value which may be tokenised into two or more sub-words.
Once the special tokens for genre are added to each sentence, we followed the same process to fine-tune mDeBERTa as before.

To evaluate the model that is trained on both NLI and genre labels, we relabelled the genre types in the dev/test set at evaluation time.
This is because dev/test dataset contains more genre types than the training dataset, and such additional genre types would not be seen at training time. 
The additional genre types in dev/test set are close enough to the training dataset genres types so that they are relabelled during evaluation using a lookup table implementing bespoke rules. 
The mapping rules we used to relabel between genres in the dev/test set to training set genres is described in Table~\ref{tab:genre-map}.
For example the \textit{Face-to-Face} genre in dev/test set is relabelled to \textit{Telephone} genre during evaluation, as they are intuitively close enough.

\begin{table}[h!]
        \footnotesize 
        \begin{tabular}{c|c} % <-- Changed to S here.
            \textbf{Dev/Test Genre} & \textbf{Train Genre} \\
            \hline
            Face-to-Face & \multirow{2}{3em}{Telephone}\\
            Telephone &\\            
            \hline
             Oxford University Press & \multirow{2}{3em}{ Slate}\\
             Letters &\\
             Slate &\\            
           \hline            
           Nine-Eleven & \multirow{2}{3em}{Government}\\
            Government &\\    
           \hline            
             Verbatim & \multirow{2}{3em}{Fiction} \\
             Fiction &\\
           \hline            
            Travel & Travel \\
        \end{tabular}
        \caption{Mapping of genre labels between training and test data}
        \label{tab:genre-map}
\end{table}

\paragraph{Combination Method}
Last but not least, we designed a final method combining some of the individual methods mentioned above. 
Even if each method provides a small uplift on Myanmar NLI performance, we believed that combining them will result in a greater uplift altogether.
Our combination method involves fine-tuning the model with a concatenated dataset of English-Myanmar cross-matched NLI examples with Genre prefixes, all from the myXNLI dataset. 
% Concatenation with ANLI or MyParaphrase data was not possible due to the lack of Genre labels in their examples. 

We discuss our evaluation results from these experiments in next section.

\subsection{Evaluation Results}

\begin{table}[h!]
        \begin{tabular}{c|c|c|c|l} % <-- Changed to S here.
            \textbf{Dataset} & \textbf{English} & \textbf{Myanmar} & \textbf{Swahili} & \textbf{Urdu}\\
            \hline
            \multicolumn{5}{c}{Cross-Lingual Transfer Baselines --- finetune on specified language data}\\
            \hline
                English & 87.24 & 75.72 (73.89) & 71.77 & 70.01\\            
            \hline
                Myanmar & 85.34 & 79.46 (78.06) & 73.15 & 73.09\\
%                Myanmar & 85.76 &  78.44 & 73.03 & 73.37\\ %tasknet
%               Myanmar & 85.76 & 79.04 & 73.49 & 73.17\\
%       (XLM-L) Myanmar & 87.64 & 79.70 & 74.83 & 75.34\\
            \hline
                Swahili & 85.02 & 76.76 (76.10) & 74.79 & 72.69 \\
            \hline
                % mdeberta-base-ur
                Urdu & 70.63 & 67.40 (66.74) & 67.86 & 66.70 \\ 
                % This doesn't look right. Could it be due to data quality? Look at another model results below.
                
                % mdeberta-v3-base-ur
                % This model is trained on data from HuggingFace instead of file extracts
                % Urdu & 73.41 & 66.52 & 67.54 & 67.06 \\
            \hline
            \multicolumn{5}{c}{Adversarial NLI Augmentation}\\
            \hline
                English + English ANLI & 87.60 & 76.32 (74.09) & 71.09 & 70.27\\
            \hline
            \multicolumn{5}{c}{Multilingual Augmentation}\\
            \hline   
                English + Myanmar & 87.74 & 80.29 (79.14) & 73.81 & 73.27\\            
            \hline  
%            mDeBERTa & English + Swahili &  &  &  & \\               
                English + Swahili & 87.14 & 77.98 (76.70) & \textbf{75.64} & 72.55\\   
            \hline           
               English + Urdu & 87.24 & 74.29 (73.77) & 74.53 & 71.27\\            
            \hline
            \multicolumn{5}{c}{Cross-matched Augmentation}\\
            \hline
                %EN-MY Cross-Matched
                \textit{en-en} + \textit{en-my} + \textit{my-en} + \textit{my-my} & 88.02 & 80.99 (79.32) & 73.41 & 74.23\\
%            mDeBERTa & English + Myanmar + Cross-Matched & 88.52 & 79.96 & 73.33 & 73.83\\
            \hline
            \multicolumn{5}{c}{Genre as Side-Input}\\
            \hline            
                English with Genre prefix & 87.84 & 75.84 (73.53) & 71.53 & 69.84\\            
%            mDeBERTa & English with Genre input & 87.62 & 74.01 & 71.15 & 69.14\\
            \hline            
                Myanmar with Genre prefix & 85.44 & 79.76 (78.76) & 73.35 & 73.73\\
%            mDeBERTa & Myanmar with Genre input & 86.18 & 79.06 & 73.27 & 73.49\\
            \hline            
                Swahili with Genre prefix & 84.87 & 77.08 (75.96) & 75.10 & 73.43\\
            \hline            
                % mdeberta-v3-base-ur-genre
                Urdu with Genre prefix & 72.03 & 67.02 (66.16) & 66.20 & 67.00\\
            \hline  
            % \multicolumn{5}{c}{NLI and Genre Multitask Fine-tuning}\\
            % \hline
            %     % mdeberta-v3-base-myxnli-mygenre_1/model_0
            %     myXNLI for NLI and Genre & 85.26 & 79.24 (77.82) & 72.63 & 72.37\\      
            % \hline            
            % \multicolumn{5}{c}{NLI and Paraphrase Multitask Fine-tuning}\\
            % \hline
            %     % mdeberta-v3-base-my-para-tn/model_0
            %    myXNLI and MyParaphase & 85.54 & 79.96 (78.38) & 73.61 & 73.23\\   % tasknet 

            % \hline            
            \multicolumn{5}{c}{Combination Method}\\
            \hline            
                 EN-MY Cross-Matched with Genre & \textbf{88.43} & \textbf{81.41 (79.96)} & 73.65 & \textbf{74.33}\\          
            \hline  
        \end{tabular}
        \caption{NLI accuracy scores for mDeBERTa$_{\text{base}}$ fine-tuned by each configuration}
        \label{tab:improved-xnli}
\end{table}

% how to read this table?
Our results in improving mDeBERTa Myanmar NLI performance are summarised in Table~\ref{tab:improved-xnli}.
As with the initial baselines, we present two results for Myanmar --- the results before the dataset revision are provided in parentheses, and the results after the revision are next to them. 
% \MDnote{230730}{Phase 1 dataset results are still the ones in parentheses, right? Mention that here. Also, best result in table again in bold.}
The baseline results for each language are at the top of this table, followed by each section representing results for each improvement method.
All experiments used the mDeBERTa base model size, fine-tuned with the corresponding datasets for a single epoch.
%In general, many approaches improved NLI performance on varying levels, with some approaches more than the other.

Overall, combining all methods works best for three of the four languages, including Myanmar (but not Swahili, where multilingual augmentation alone is best).  Fine-tuning on own language generally produces the best cross-lingual transfer baseline, except in the odd case of Urdu: fine-tuning on Urdu dramatically worsens performance on all languages, including itself, while fine-tuning on Myanmar gives the best results for Urdu.  As before, Myanmar has the best results of the low-resource languages under these extensions as well, again supporting its quality.

Considering then the improvements from including data augmentation relative to the best fine-tuned models, these range from around $1$ percentage point for Swahili (74.79\% vs 75.64\%) to around $2$ for Myanmar (79.46\% vs 81.41\%).  This gap is around the same as the improvement gained by improving our Myanmar dataset quality, which as in Table~\ref{tab:baseline-myxnli} ranges up to 2 percentage points as well and even slightly higher in some cases.  Improving dataset quality, then, gives benefits at least equivalent to tinkering with a range of improved training techniques.

\section{Conclusion and Future Work}

% What did we cover?
In this paper, we explored Natural Language Inference (NLI) in Myanmar language as a proxy topic for a broader challenge in low-resource Natural Language Understanding (NLU).
Specifically, we have addressed the questions guiding our research as follows.

\paragraph{
What are the challenges and solutions in building a dataset in a low-resource language such as Myanmar?
}

Through our efforts building the myXNLI dataset, we have uncovered several types of challenges and solutions that may be generally present in building low-resource datasets. 
Extending existing multilingual datasets facilitates building low-resource datasets by enabling the reuse of existing data sources, annotations and parallel data.
While it is possible to build such datasets mainly based on volunteer participation, we found that this must be enabled by collaborative tools (e.g. github, shared drives) and processes to encourage participation (e.g. regular check-ins and leaderboards).
Using local translators or annotators may lead to sub-optimal outcomes due to limited bilingual skills and cultural context, but recruiting skilled volunteers may pose coordination and participation challenges across geo-locations.
%\MDnote{230801}{Added following sentence.}
We found that a two-stage process in this kind of low-resource context, first using local translators and annotators and then followed by skilled workers for reviewing, leads to an improvement in dataset quality that is reflected in model performance on tasks; in our case, for NLI, this was up to 2 percentage points in accuracy.  The magnitude of this improvement is similar to that of tinkering with several methods for improving training.

Regardless of skills and background of the participants, one may still encounter translation and transliteration challenges in relation to the task and language at hand, as we found with arbitrary translations in Myanmar leading to NLI errors.
Therefore, we found the importance of establishing clear and comprehensive translation annotation guidelines as early as possible, and developing tools to enforce them (e.g. shared dictionaries, validation scripts).

\paragraph{
What are the performance and limitations of recent language models in Myanmar Language?
}  

To answer this question, we consider NLI as a key task representing broader NLU challenges in Myanmar and provide our analysis based on the Myanmar NLI results.
Our baseline results of selected recent models on myXNLI benchmark suggests that state-of-the-art language models have the potential to work well in Myanmar language but they are highly under-tuned towards the language mainly due to the lack of datasets.
In particular, the \textit{translate-train} results of multilingual model mDeBERTa suggests that fine-tuning multilingual models with in-language datasets provides the best performance for Myanmar.
On the other hand, the \textit{translate-test} results of DeBERTaV3 also showed that using machine-translation together with high-resource monolingual models is a reasonable alternative for Myanmar in the absence of task-specific multilingual or Myanmar-only models.
We also observed considerable cross-lingual transfer from English to Myanmar, such that English data may be used to fine-tune multilingual models and used for Myanmar when other options are not available.  
While there were not many Myanmar-only monolingual models to provide a comparison, we found that multilingual models perform much better than Myanmar-only models at least for the NLI task. 
Nevertheless, the multilingual models still have limitations as indicated by our error analysis of the best-performing mDeBERTa model (Appendix \ref{secA1}).
% In particular, we found that multilingual models still have limited understanding of Myanmar words which could be due to a limitation in pretraining.

\paragraph{
What are some strategies to improve model performance on Myanmar?
}

%Through our experiments in Chapter~\ref{chap:methods}, we found several strategies to improve the model performance in Myanmar.
We observed a strong cross-lingual transfer between the high-resource language English and the low-resource language Myanmar, and this led to multiple approaches that combine English and Myanmar data together. 
Fine-tuning on English and Myanmar together on the same task improved the Myanmar performance, and it is possible that adding more languages can lead to further uplifts.
Cross-matching English and Myanmar examples in the same dataset also improved  Myanmar performance. 
In the absence of Myanmar data entirely, simply training more on high-resource datasets such as English ANLI may also lead to improvements by means of cross-lingual transfer. 
In a different strategy, we also showed that it is possible to exploit existing metadata such Genre to improve classification tasks such as NLI. 
% Additionally, multitask fine-tuning with different Myanmar task datasets could be considered as a potential strategy if there is adequate data in the supplementary datasets.
Last but not least, we found that it is possible to combine multiple compatible strategies together as seen in our combination method of fine-tuning on cross-matched English-Myanmar data with Genre prefixes to create a larger overall effect.

\paragraph{
Can the strategies for Myanmar be used for other low-resource languages? 
}

As we evaluated several strategies to improve Myanmar NLI performance, we also did the same for our reference low-resource languages, Swahili and Urdu.
We believe our methods are language-agnostic, as we did not address Myanmar-specific problems such as the existence of non-Unicode content in pretrained data.
Our findings confirmed that there is some cross-lingual transfer from English to Swahili and Urdu, therefore English data can be leveraged to improve them.
Training English data together with Swahili or Urdu datasets uplifted their respective performances, but the results vary between languages. 
For Urdu, fine-tuning on English alone provides a better result than fine-tuning only on Urdu.
For Swahili, training just on English XNLI is just as good as training on combined English XNLI and ANLI data.
We also confirmed that using Genre prefixes uplifted performance for both languages, suggesting that other metadata may be leveraged in similar ways.
Although we did not evaluate cross-matching between English and Swahili or Urdu, our results for Myanmar suggests that they could be applied the same.
Overall, we present our view that the strategies developed for Myanmar can be, in fact used  for other low-resource languages in general.
% We encourage the research community in other low-resource languages to explore these strategies in their languages respectively.

\paragraph{Future work}
Natural Language Understanding for low-resource languages remains a challenging problem, despite the visible uplift in high-resource languages achieved by recent language models.
Our efforts to benchmark and improve NLI for Myanmar language rather suggests that there is much work left to be done in the domain of low-resource languages.
In this spirit, we aim to explore the following few areas as our future work. 
In the short term, we aim to include myXNLI in cross-lingual benchmarks such as XTREME \citep{ruder-etal-2021-xtreme} to maximise its impact.
There is also scope to evaluate other transformer architectures such as mT5 \citep{mt5-xue-etal-2021} and Aya \citep{ustun2024aya} against the myXNLI benchmark.
Creating additional synthetic training data for Myanmar NLI using language models, with techniques similar to PromDA \cite{promda_wang-etal-2022} is another approach yet to be explored.
For the longer term, architectural innovations in language models provide a promising approach for the low-resource challenges, as seen in the uplift between XLM-R and mDeBERTa performance on Myanmar. 
Last but not least, leveraging our experience, technology and community networks established in building myXNLI, we will endeavour to develop more Myanmar/multilingual datasets to create similar profound effects on the low-resource NLP community. 
%In particular, we aim to construct supplementary Myanmar datasets for MASSIVE \citep{fitzgerald-etal-2023-massive} to improve semantic parsing performance for Myanmar, which can ultimately uplift the quality of virtual assistants for Myanmar language users.

%\section{Guidance: Conclusion}\label{sec13}

%Conclusions may be used to restate your hypothesis or research question, restate your major findings, explain the relevance and the added value of your work, highlight any limitations of your study, describe future directions for research and recommendations. 

%In some disciplines use of Discussion or 'Conclusion' is interchangeable. It is not mandatory to use both. Please refer to Journal-level guidance for any specific requirements. 

%\backmatter

%\bmhead{Supplementary information}

%If your article has accompanying supplementary file/s please state so here. 

%Authors reporting data from electrophoretic gels and blots should supply the full unprocessed scans for key as part of their Supplementary information. This may be requested by the editorial team/s if it is missing.

%Please refer to Journal-level guidance for any specific requirements.

\bmhead{Acknowledgments}
% The underlying research for this paper was done as a Masters thesis by Aung Kyaw Htet and supervised by Mark Dras at Macquarie University.
The initial translation guidelines and efforts were contributed by a team of Myanmar NLP researchers led by Win Pa Pa, and further contributed by several volunteers across different geo-locations. The names of all translators are available online.

%Acknowledgments are not compulsory. Where included they should be brief. Grant or contribution numbers may be acknowledged.

%Please refer to Journal-level guidance for any specific requirements.

\section*{Declarations}

%Some journals require declarations to be submitted in a standardised format. Please check the Instructions for Authors of the journal to which you are submitting to see if you need to complete this section. If yes, your manuscript must contain the following sections under the heading `Declarations':

\bmhead{Availability of data and materials}
The Myanmar XNLI dataset and a corresponding fine-tuned model is available on Github and HuggingFace.
\begin{itemize}
\item Dataset Repository \url{github.com/akhtet/myXNLI}
\item Dataset for Fine-tuning \url{huggingface.co/datasets/akhtet/myanmar-xnli}
\item Fine-tuned Model \url{huggingface.co/akhtet/mDeBERTa-v3-base-myanmar-xnli}
\end{itemize}

% \bmhead{Conflict of interest} None

\bmhead{Funding} Translation efforts for the Myanmar XNLI dataset beyond volunteer contributions were funded by Macquarie University.

%\begin{itemize}
%\item Conflict of interest/Competing interests (check journal-specific guidelines for which heading to use)
%\item Ethics approval 
%\item Consent for publication
%\item Availability of data and materials
%\item Authors' contributions
%\end{itemize}

%\noindent
%If any of the sections are not relevant to your manuscript, please include the heading and write `Not applicable' for that section. 

%\bigskip

%\bigskip\noindent
%Springer journals and proceedings: \url{https://www.springer.com/gp/editorial-policies}

%\bigskip\noindent
%Nature Portfolio journals: \url{https://www.nature.com/nature-research/editorial-policies}

%\bigskip\noindent
%\textit{Scientific Reports}: \url{https://www.nature.com/srep/journal-policies/editorial-policies}

%\bigskip\noindent
%BMC journals: \url{https://www.biomedcentral.com/getpublished/editorial-policies}
%\end{flushleft}
\newpage
\begin{appendices}

\section{Error Analysis on Myanmar NLI Results}\label{secA1}


In this section, we present our analysis of some outputs from the models. 
For our analysis, we compared the outputs from baseline Myanmar model to the outputs of improved Myanmar models.
We found that some NLI examples which were previously incorrectly predicted in the baseline were corrected by the improved methods.
On the other hand, a few examples which were correctly predicted in the baseline became incorrect after applying certain methods.

\subsection{Effects of Translation Revision}

\begin{figure}[h]  
	\begin{center}
      \includegraphics[width=1\linewidth]{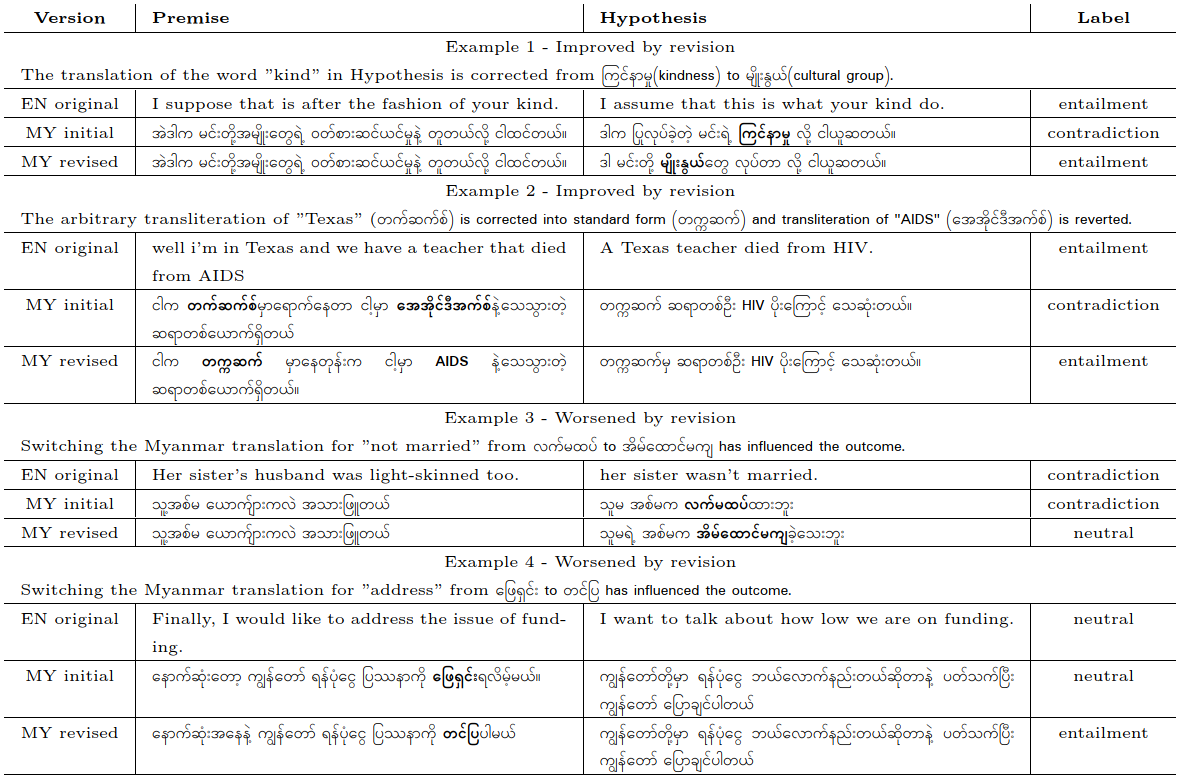}
    \end{center}
	\caption{Examples for positive and negative effects of translation revision on myXNLI}
	\label{fig:revision-effects-myxnli}
\end{figure}

As seen in our baseline and improved Myanmar NLI results, we obtained higher scores after switching to the revised dataset. 
We explored the effects of translation revision in detail by analysing the results before and after.
We used the mDeBERTa model fine-tuned on Myanmar only (baseline model) to evaluate on both the initial and revised myXNLI dataset.
We found that after the revision, some predictions were corrected while some predictions were misjudged.
More precisely, with the revised dataset, 202 predictions which were previously incorrect became correct and 132 predictions which were previously correct became incorrect, resulting in the overall improvement in accuracy.
From the results, we take 2 examples each from correct and incorrect examples, and show them in Figure \ref{fig:revision-effects-myxnli}.

\paragraph{Improved Examples}
We observed that correcting wrong word-senses or phrases during translation revision led to correct predictions (Example 1).
Also, standardising arbitrary transliterations or avoiding transliteration completely led to correct predictions (Example 2).

\paragraph{Worsened Examples}
On the other hand, simply rewriting some Myanmar words with their synonyms or more appropriate terms with similar meanings may lead to incorrect results (Examples 3 and 4).
This suggests that the meaning of some Myanmar words are less well-understood by the model than some other words, possibly due to the effects of pre-training. 
In addition, the correct prediction for Example 4 could be argued as \textit{entailment} by some individuals.
In fact, in the original XNLI data, the collective labels for this example is (4) neutral and (1) entailment for this, assigning neutral as the gold label based on majority vote.

Overall, we found that improving the translation and standardising transliteration have positive effects on the Myanmar NLI results, although the usage of some Myanmar words may inadvertently confuse the model as a minor side-effect.

\subsection{Effects of Genre as Side-Input}

Our method to use Genre as side-input obtained a small but consistent improvement across all languages.
To understand the effects of using Genre, we compare the results of a baseline mDeBERTa model fine-tuned on Myanmar only and a mDeBERTa model fine-tuned on Myanmar with Genre prefixes.
In comparison, we found that by using the latter, 147 examples become corrected, and 132 examples become misjudged, resulting in an overall improvement.
To explain the outputs, it was necessary examine the effects of Genre on the attention weights of our model outputs. 
We used transformer-interpret library\footnote{\url{https://github.com/cdpierse/transformers-interpret}} to depict the attribution for each prediction. 
In the following figures generated by this library, the Myanmar phrases high-lighted in Green contribute positively to the predicted labels (i.e. tokens influencing towards this prediction) while those in Red contribute negatively (influencing against this decision).  
The intensity of colors also indicate their levels of influence in doing so.

\begin{figure}[h]  
	\begin{center}
      \includegraphics[width=0.9\linewidth]{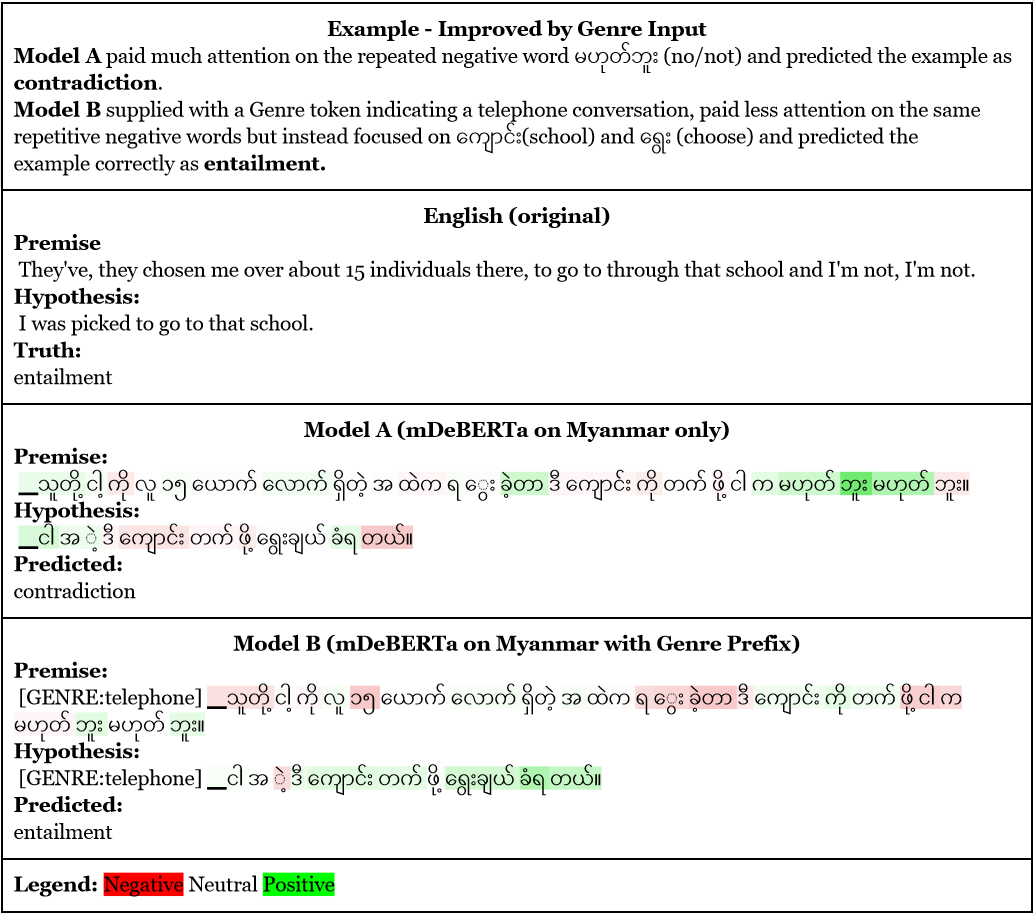}
    \end{center}
	\caption{An example of positive effect by using Genre as Side-Input}
	\label{fig:genre-effect-positive}
\end{figure}

\paragraph{Improved by Genre Input}
In Figure \ref{fig:genre-effect-positive}, we provide an example of a NLI task that was incorrectly predicted by baseline model, but correctly predicted by the genre-aware model, having Genre as side-input (as Prefixes).
Our intuitive explanation is that in telephone conversations, it is common to find repeated short utterances such as ``Yes" and ``No", and they should be treated differently (perhaps more lightly) than the occurrence of similar words in more written-style genres.
The baseline model (model A) was not aware of the nature of the input as telephone conversation.
In contrast, the genre-aware model (model B) provided with the genre prefix token focuses instead on other important words in the task and correctly predicted the label.
%\MDnote{230730}{Explain briefly what the reader is seeing in the figure in terms of the colour-coding.}

\begin{figure}[h]  
	\begin{center}
      \includegraphics[width=0.9\linewidth]{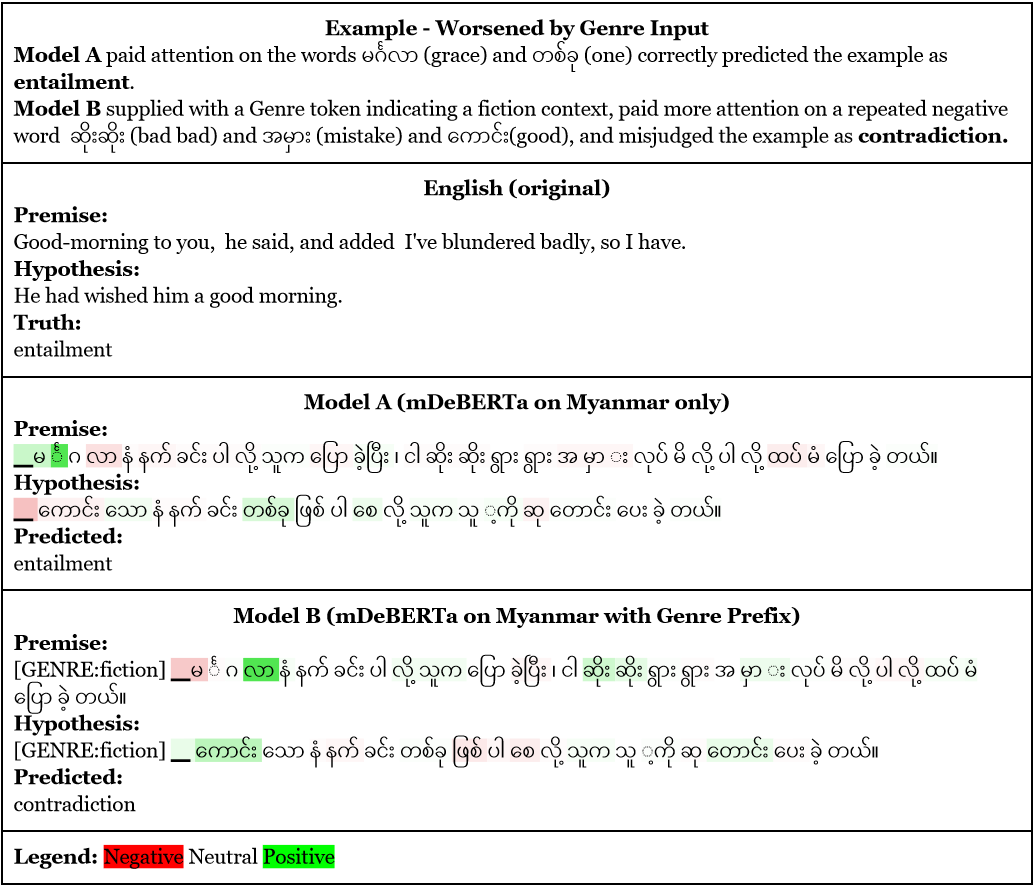}
    \end{center}
	\caption{An example of negative effect by using Genre as Side-Input}
	\label{fig:genre-effect-negative}
\end{figure}

\paragraph{Worsened by Genre Input}
We also present an example of a previously correct example by the baseline model now becoming incorrect after using the Genre input in Figure~\ref{fig:genre-effect-negative}.
Continuing from our general assumption about the nature of the telephone conversations, it could be argued that written-style genres should pay more attention on repetitive and negative words, unlike in spoken-style genres. 
In this example, the genre-aware model has paid much attention on the repeated positive and negative words, leading to an incorrect prediction. 
One explanation could be that the genre-aware model had over-generalised the learnings between spoken-style and written-style genres too far.

\subsection{Effects of space characters in Myanmar}

\begin{figure}[h]  
	\begin{center}
      \includegraphics[width=0.9\linewidth]{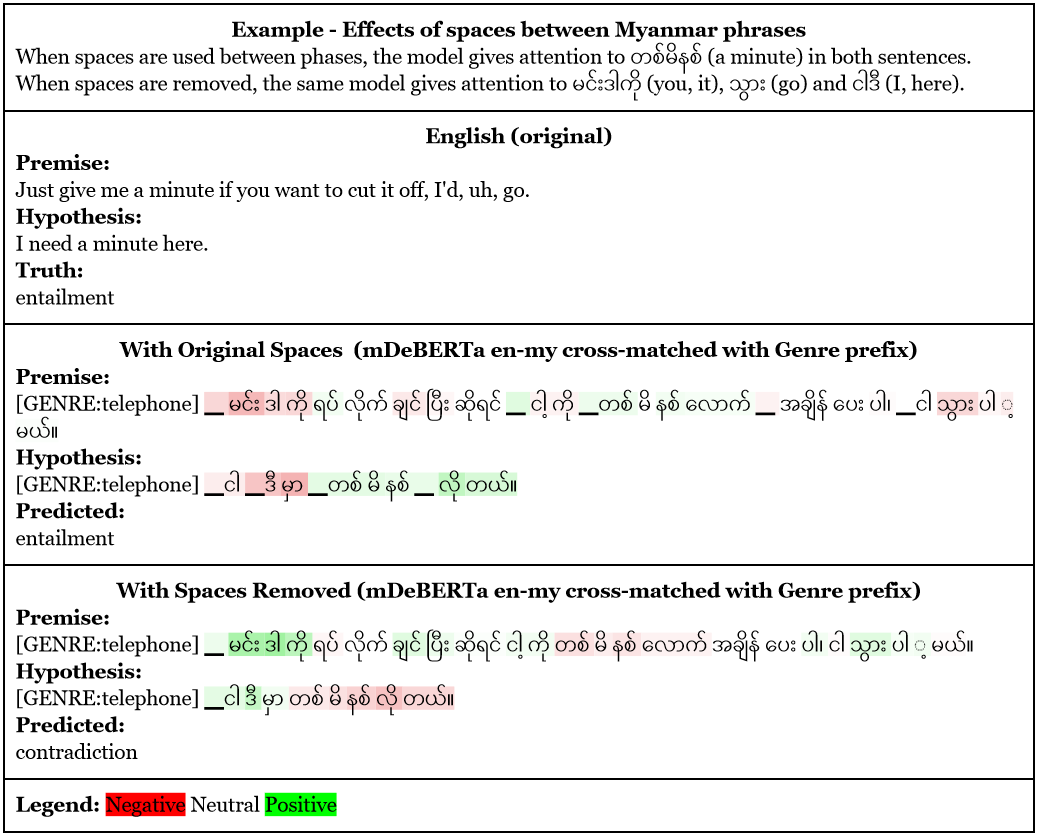}
    \end{center}
	\caption{An example of negative effect by removing spaces in input}
	\label{fig:space-effect-negative}
\end{figure}

Space characters are optional in Myanmar script and only used optionally between phrases for readability reasons.
Even when spaces are used, their placements are rather arbitrary and depends on the author.
As such, one might conclude that spaces are not considered as important tokens by models.
On the contrary, we found that removing spaces can have some effects on the NLI performance.
We evaluated our best model (mDeBERTa en-my cross-matched with Genre prefix) with the test set with spaces removed and compared to the results using the original test set.
This model was fine-tuned with the training data that includes spaces.
When the spaces are removed from the test set, 138 examples which were correct became incorrect, while 106 incorrect examples became correct.
Figure \ref{fig:space-effect-negative} provides one example which became incorrect once the spaces are removed from the input.
We also found opposite but fewer examples where removing spaces in the input leads to the correct predictions.
This suggests that some spaces may be rather confusing the model.
The mismatch between training and evaluation data in terms of spaces may have also led to the overall negative effect on accuracy.
We leave the comprehensive analysis on the effects of spaces for future work.
%\MDnote{230730}{Maybe also comment }

\subsection{Remaining and persistent errors}

We also explored the remaining errors which are persistent through multiple models in our experiments.
Rather than comparing every model, we explored the common errors between the baseline model (Myanmar only) and best model (en-my cross-matched with Genre prefix).

\begin{figure}[h]  
	\begin{center}   \includegraphics[width=1\linewidth]{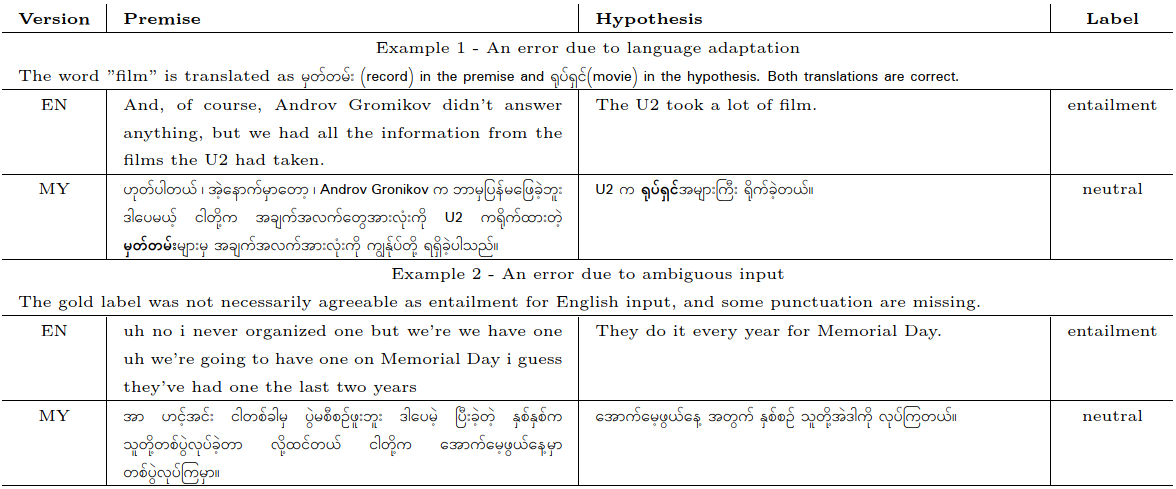}
    \end{center}
	\caption{Examples of language and input error types}
	\label{fig:language-input-error-types}
\end{figure}

\begin{figure}[h]  
	\begin{center}
      \includegraphics[width=0.9\linewidth,height=0.9\textwidth]{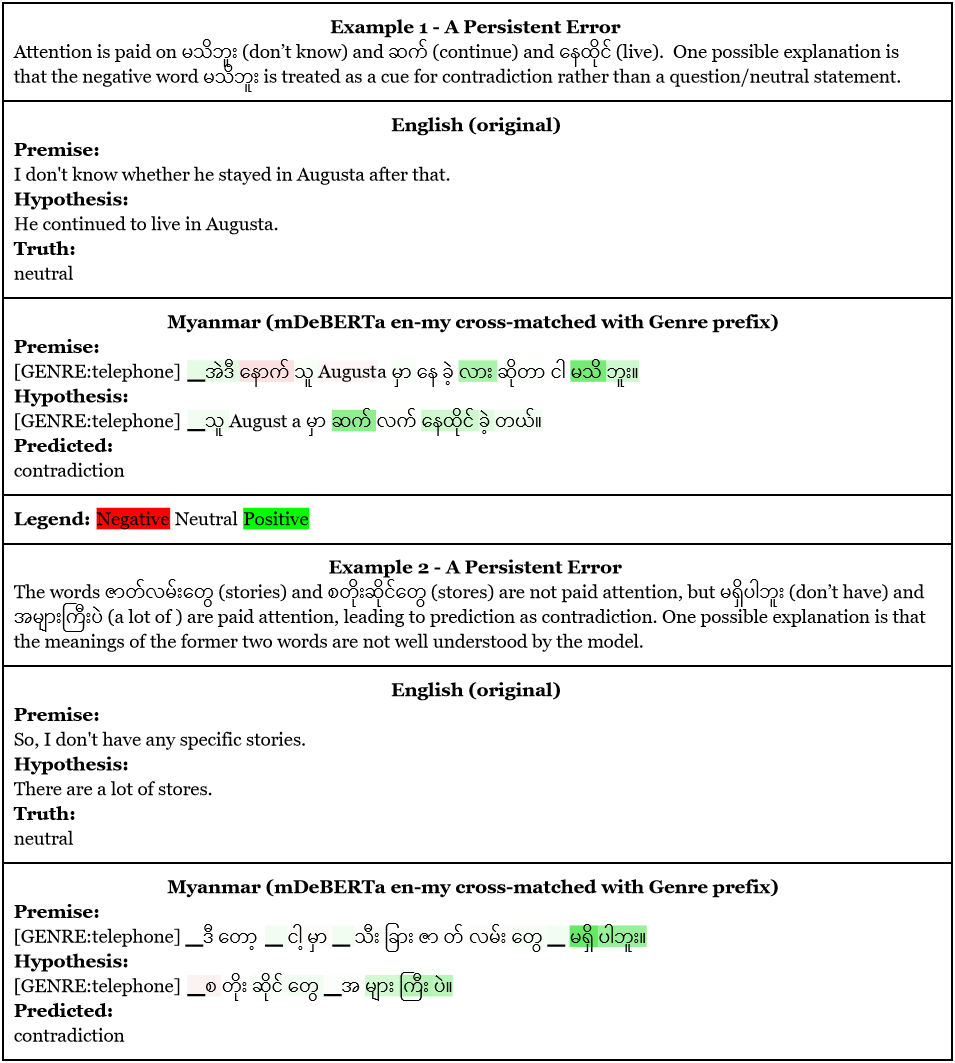}
    \end{center}
	\caption{Examples of persistent errors after our best Myanmar model}
	\label{fig:persistant-errors}
\end{figure}

To get a statistical view of these errors, we sampled 100 NLI pairs from the test set which are misclassified by both models.
For most errors in these samples, both models consistently predicted the same label although they disagree with the gold label.
We also manually analysed each error in the samples and categorised them into the following error categories: \textit{Translator, Language, Input} and \textit{Model}.
The description of each error category along with its corresponding count within the sample set is described in Table \ref{tab:nli-error-stats}.
Most of these errors involve sentences with bad translations or transliterations where a better Myanmar representation could have avoided the issue.
However, there are also errors caused by the challenges in adapting English into Myanmar words within limited context in each sentence regardless of the translation skills.
We also found that some errors are rather caused by the English input sentences, which are ambiguous, illegible, or the gold label is not necessarily agreeable.
We provide examples of language and input errors in Figure \ref{fig:language-input-error-types}.
However some errors appear to be genuine prediction errors which cannot be classified as caused by translator, language adaption or input.
We provide two examples of such genuine model errors in Figure \ref{fig:persistant-errors} along with possible explanations.
%\MDnote{230730}{You've repeated this: there's only one figure / error.} a
As with XNLI tasks generally, positive and negative words tend to heavily influence the predictions, but our analysis also suggested that a lack of understanding of surrounding words also contributed towards the errors.

\begin{table}[h!]
\footnotesize
\begin{tabular}{p{0.1\linewidth} | p{0.7\linewidth} | p{0.1\linewidth}}
    \textbf{Category} & \textbf{Description} & \textbf{Count}\\
    \hline
    Translator & Error possibly due to an improper translation or transliteration. & 39\\
    \hline
    Language & Error possibly due to adaption between English and Myanmar words. &  16\\
    \hline
    Input &  Error possibly due to ambiguity in the English source input itself. &  22\\
    \hline
    Model & Genuine error by the model or cause unknown otherwise. &  23\\
    \end{tabular}
    \caption{Error types and distributions from error samples by two selected models}
    \label{tab:nli-error-stats}
\end{table} 

\subsection{Unexplored areas in error analysis}

\paragraph{Out of domain generalisation}

So far, we have found some evidence that training models with genre information can improve the results, as indicated by the example in Figure \ref{fig:genre-effect-positive} and the accuracy results in Table \ref{tab:improved-xnli}.
However, we have not yet explored the opposite i.e. how the models behave when provided with the wrong genre information, or when there are significant variations between the domain of the training data and the test data.
More generally, we have not yet explored how the models behave when trained on a particular domain or genre but used in a different context, similar to challenges often encountered in real-world applications.
Since the myXNLI dataset contains examples across different domains or genres, it is possible to explore such issues in the future by creating cross-domain evaluations in the interest of creating more robust models.

\paragraph{Myanmar-specific issues}

Our analysis so far explored language agnostic issues (except the effect of space character positioning).
We leave it to future work to explore more Myanmar-specific issues, such as the effects of morphological variations in Myanmar input data, given the rich morphology of the language.
In particular, one could explore if morphological variants are still recognised during tokenisation by mDeBERTa and generating correct inference results.
It is also not yet known how much of the Myanmar input is treated as unknown tokens.
This could potentially explore additional pre-training for model and extending its vocabulary in Myanmar towards creating better results.



\end{appendices}

\newpage
\bibliography{sn-bibliography}% common bib file

\end{document}